\documentclass[10pt,journal,compsoc]{IEEEtran}
\usepackage{enumitem}
\usepackage{xr}
\usepackage[normalem]{ulem}
\usepackage{subfigure}
\usepackage{graphicx}
\usepackage{comment}
\usepackage{amsmath}
\usepackage[table]{xcolor}
\usepackage{amsfonts}
\usepackage{amssymb}
\usepackage{algorithm}
\usepackage{algpseudocode}
\usepackage{booktabs}
\usepackage{longtable}
\usepackage{xr}
\usepackage{tikz}
\usetikzlibrary[automata, positioning]
\usepackage{hyperref}
\hypersetup{colorlinks=false}
\usepackage[flushleft]{threeparttable}
\usepackage{multirow}
\usepackage{caption}
\usepackage{slashbox}
\usepackage{xspace}
\usepackage{verbatim,soul}
\usepackage{rotating}

\makeatletter
\let\pgfmathrandomX=\pgfmathrandom@
\usepackage{pgfplotstable}
\let\pgfmathrandom@=\pgfmathrandomX
\makeatother

\makeatletter
\def\@xfootnote[#1]{%
  \protected@xdef\@thefnmark{#1}%
  \@footnotemark\@footnotetext}
\makeatother

\newcommand{\minus}{\scalebox{0.6}{$-$}}
\newcommand{\plus}{\scalebox{0.6}{$+$}}

\newcommand{\CI}{\mbox{$\mathop{\mathtt{CI}}\limits$}\xspace}
\newcommand{\rCI}{\mbox{$\mathop{\mathtt{sCI}}\limits$}\xspace}

\newcommand{\BMTMKL}{\mbox{$\mathop{\mathtt{BMTMKL}}\limits$}\xspace}
\newcommand{\MTCRP}{\mbox{$\mathop{\mathtt{pLETORg}}\limits$}\xspace}
\newcommand{\KRL}{\mbox{$\mathop{\mathtt{KRL}}\limits$}\xspace}

\newcommand{\LETOR}{\mbox{$\mathop{\mathtt{LETOR}}\limits$}\xspace}

\newcommand{\drug}{\mbox{$\mathop{d}\limits$}\xspace}
\newcommand{\cellline}{\mbox{$\mathop{\mathcal{C}}\limits$}\xspace}

\newcommand{\efdiff}{\mbox{$\mathop{\Delta\text{e\%}}\limits$}\xspace}
\newcommand{\prdiff}{\mbox{$\mathop{\Delta\text{r\%}}\limits$}\xspace}
\newcommand{\simt}{\mbox{$\mathop{\mathtt{TAN}_{\text{AF}}}\limits$}\xspace}
\newcommand{\siml}{\mbox{$\mathop{\mathtt{COS}_{\text{L}}}\limits$}\xspace}

\newcommand{\etal}{\mbox{\emph{et al.}}\xspace}

\newcommand{\holdoutsize}{\mbox{$\mathop{N_{\text{new}}}\limits$}\xspace}

\newcommand{\gl}{\mbox{{corr(GE, LV)}}\xspace}
\newcommand{\gd}{\mbox{{corr(GE, DS)}}\xspace}
\newcommand{\ld}{\mbox{{corr(LV, DS)}}\xspace}

\newcommand{\COS}{\mbox{$\mathop{\mathtt{COS}}\limits$}\xspace}
\newcommand{\RBF}{\mbox{$\mathop{\mathtt{RBF}}\limits$}\xspace}
\newcommand{\LIN}{\mbox{$\mathop{\mathtt{LIN}}\limits$}\xspace}

\newcommand{\subparagraph}{}
\DeclareRobustCommand{\hlgreen}[1]{{\sethlcolor{green}\hl{#1}}}

\renewcommand\hl[1]{#1} 
\renewcommand\hlgreen[1]{#1} 

\ifCLASSOPTIONcompsoc
  \usepackage[nocompress]{cite}
\else
  \usepackage{cite}
\fi

\graphicspath{{./plot/}{./figures/}}

\begin{document}

\title{Drug Selection\\via Joint Push and Learning to Rank}

%

\author{Yicheng~He$^*$,
        Junfeng~Liu$^*$,
        and~Xia~Ning$^\ddagger$
\IEEEcompsocitemizethanks{
\IEEEcompsocthanksitem
Y.~He$^*$ and J.~Liu$^*$ have equal contributions to this work and are
co-first authors.
\IEEEcompsocthanksitem 
Y.~He, J.~Liu and X.~Ning are with the Department of Computer and Information Science, Indiana University - Purdue University Indianapolis.
%
%
\IEEEcompsocthanksitem 
X.~Ning is with the Center for Computational Biology and Bioinformatics, Indiana University School of Medicine.\protect\\
Address: 410 West 10th St., HITS 5011, Indianapolis, IN 46202, USA \protect\\
$^\ddagger$E-mail: xning@iupui.edu
}
}

\markboth{IEEE/ACM Transactions on Computational Biology and Bioinformatics}%
{}

\IEEEtitleabstractindextext{%
  \begin{abstract}
    Selecting the right drugs for the right patients is a primary goal of precision
    medicine.
    In this manuscript, we consider the problem of cancer drug selection in a learning-to-rank
    framework. 
    We have formulated the cancer drug selection problem as to 
    accurately predicting 1). the ranking positions of sensitive drugs
    and 2). the ranking orders among sensitive drugs in cancer cell lines based on
    their responses to cancer drugs. 
    We have developed a new learning-to-rank method, denoted as \MTCRP, that predicts
    drug ranking structures in each cell line via 
    using drug latent vectors and cell line latent vectors.
    The \MTCRP method learns such latent vectors through explicitly enforcing that,
    in the drug ranking list of each cell line, 
    the sensitive drugs are pushed above insensitive drugs, and
    meanwhile the ranking orders among sensitive drugs are correct. 
    %
    Genomics information on cell lines is leveraged in learning the latent vectors. 
    Our experimental results on a benchmark cell line-drug response dataset demonstrate that
    the new \MTCRP significantly outperforms the state-of-the-art method in prioritizing new
    sensitive drugs.
\end{abstract}
\begin{IEEEkeywords}
Drug Selection, Learning to Rank
\end{IEEEkeywords}}

\maketitle

\IEEEdisplaynontitleabstractindextext

%
\IEEEpeerreviewmaketitle

\IEEEraisesectionheading{\section{Introduction}\label{sec:intro}}

\IEEEPARstart{S}{electing} the right drugs for the right patients is a primary goal
of precision medicine~\cite{Ashley2016}.
An appealing option for precision cancer drug selection is via the
pan-cancer scheme~\cite{Omberg2013} that examines various cancer types together.  
%
The landscape of cancer genomics reveals that various cancer types share 
driving mutagenesis mechanisms and corresponding molecular signaling 
pathways in several core cellular processes~\cite{Vogelstein2013}. 
This finding has motivated the most recent clinical trials (e.g.,
the Molecular Analysis for Therapy Choice Trial at National Cancer
Institute\footnote{https://www.cancer.gov/about-cancer/treatment/clinical-trials/nci-supported/nci-match})
to identify common targets for patients of various
cancer types and to prescribe same drug therapy to such patients.
Such pan-cancer scheme
is also well supported by the strong pan-cancer
mutations~\cite{Kandoth2013} and copy number variation~\cite{Zack2013} patterns observed from 
The Cancer Genomics Atlas\footnote{https://cancergenome.nih.gov/} project.
The above pan-cancer evidence from theories and practices lays the foundation for
joint analysis of multiple cancer cell lines and their drug responses to prioritize
and select sensitive cancer drugs. 

%
Another appealing option for precision cancer drug selection 
is via the popular off-label drug use~\cite{Conti2013} (i.e., the use of drugs
for unapproved therapeutic indications~\cite{Stafford2008}).
This is due to the fact that some aggressive cancer types have very limited existing
therapeutic options, while conventional drug development for those cancers, and also
in general, has been extremely time-consuming, costly and risky~\cite{DiMasi2003}. 
However, a key challenge for off-label drug use is the lack of knowledge base
of preclinical and clinical evidence, hence, the guidance for drug selection
in practice~\cite{Poole2004}.
In this manuscript, we present a new computational cancer drug selection method --
joint \ul{\textbf{p}}ush and \ul{\textbf{LE}}arning \ul{\textbf{TO}} \ul{\textbf{R}}ank with
\ul{\textbf{g}}enomics regularization (\MTCRP). 
In \MTCRP, we formulate the problem of drug selection based on cell line responses
as a learning-to-rank~\cite{Liu2009} problem, that is, we aim
to produce accurate drug orderings (in terms of drug sensitivity)
in each cell line via learning, and thus
prioritize sensitive drugs in each cell line.
This corresponds to the application scenario in which drugs need to be
prioritized and selected to treat a given cell line/patient. 
Drug sensitivity here represents the capacity of drugs
for reduction in cancer cell proliferation. Cell line responses to drugs reflect
drug sensitivities on the cell lines, and thus, we use the concepts of
drug sensitivity and cell line response in this manuscript exchangeably. 
%

To induce correct ordering of drugs in each cell line in terms of drug sensitivity, 
for each involved drug and cell line, in \MTCRP, we learn a latent 
vector and score drugs in each cell line using drug latent vectors and the
corresponding cell line latent vector. 
%
%
We learn such latent vectors through explicitly enforcing and optimizing that,
in the drug ranking list of each cell line, 
the sensitive drugs are pushed above insensitive drugs, and
meanwhile the ranking orders among sensitive drugs are correct,
where the ranking position of a drug in a cell line is determined by the
drug latent vector and cell line latent vector. 
We simultaneously learn from all the cell lines and their drug ranking structures. 
%
%
In this way, the structural information of all
the cell lines can be transferred across and leveraged during the learning process.
We also use genomics information on cell lines to regularize the latent vectors 
in learning to rank.
Fig.~\ref{fig:scheme} demonstrates the overall scheme of the \MTCRP method.
\begin{figure}[h!]
  \begin{center}
  \fbox{
    \scalebox{1.1}{
      \input{figures/scheme.tex}
    }
  }
  \vspace{5pt}
  \caption{\MTCRP scheme overview}
  \label{fig:scheme}
  \end{center}
  \vspace{-15pt}
\end{figure}
The new \MTCRP is significantly different from the existing computational
drug selection methods. 
Current computational efforts for precision cancer drug selection~\cite{Niz2016}
are primarily focused on using regression methods (e.g., 
kernel based methods~\cite{Gonen2012, He2018}, 
\hl{matrix factorization}~\cite{Zheng2013, Liu2016}, 
\hl{network-based similarity aggregation}~\cite{Zhang2015predicting})
to predict drug sensitivities (e.g., in GI$_{50}$~\footnote{https://dtp.cancer.gov/databases\_tools/docs/compare/compare\\\_methodology.htm},
IC$_{50}$~\footnote{https://www.ncbi.nlm.nih.gov/books/NBK91994/})
numerically, and selecting drugs with optimal sensitivities in each cell
line~\cite{Costello2014}.
For example, in Menden \etal~\cite{Menden2013},
cell line features (e.g., sequence variation, copy number variation) and drug
features (e.g., physicochemical properties) are jointly used to train a neural
network that predicts drug sensitivities in IC$_{50}$ values. 
\hl{
  Drug selection has also been formulated as a classification problem. 
  For example, in Stanfield {\etal}{~\cite{Stanfield2017}}, cell line profiles, and drug
  sensitive and resistant profiles are generated from heterogeneous
  cell line-protein-drug interaction networks to score drugs in cell lines. The
  scores are compared with a threshold 
  to classify whether the drugs are sensitive in the cell lines. 
}
%
%
%
Another focus of the existing methods is on effectively using genomics information on
cell lines and features on drugs to improve regression~\cite{Liu2016, Gonen2012}.
%
For example, in Ammad-ud-din \etal~\cite{Ammad2014}, a kernel is constructed on each type
of drug and cell line features to measure their respective similarities,
and drug sensitivity is predicted from the combination of projected drug kernels and cell
line kernels.

\begin{figure}[!h]
\vspace{-10pt}
  \begin{center}
    \input{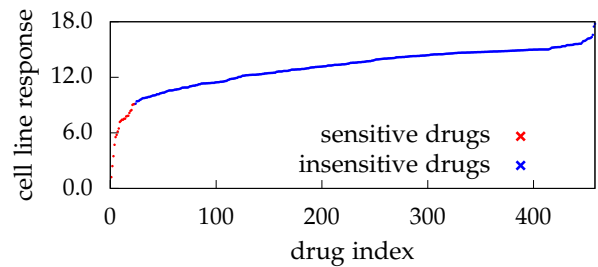}
    \caption{Exemplar cell line response score distribution}
    \label{fig:CL-LS123-distribution}
  \end{center}
  \vspace{-10pt}
\end{figure}

The existing regression based methods for drug selection may suffer from the fact that
the regression accuracy is largely affected by insensitive drugs, 
and therefore, accurate drug sensitivity regression does not necessarily lead to
accurate drug selection (prioritization).
This is because in regression models, in order to achieve small regression errors,
the majority of drug response values in a cell line needs to be fit well.
However, when insensitive drugs constitute the majority in each cell line,
which is becoming common as the advanced technologies are enabling screenings over 
large collections of small molecules (e.g., in the
Library of Integrated Network-Based Cellular
Signatures (LINCS)~\footnote{http://www.lincsproject.org/}), 
it is very likely that the regression sacrifices its
accuracies on a very few but sensitive drugs in order to achieve better accuracies
on the majority insensitive drugs, and thus smaller total errors on all drugs overall.
This situation is even more likely when the cell line response values on
sensitive drugs follow a very different distribution, and thus appear like outliers~\cite{Speyer2017}, 
than that from insensitive drugs, which is also very often the case.
Fig.~\ref{fig:CL-LS123-distribution} presents a typical distribution of
cell line (LS123 from Cancer Therapeutics Response
Portal (CTRP v2)~\footnote{https://portals.broadinstitute.org/ctrp/}) responses to drugs.
In Fig.~\ref{fig:CL-LS123-distribution}, lower cell line response scores indicate higher
drug sensitivities. It is clear in Fig.~\ref{fig:CL-LS123-distribution} that
top most sensitive drugs (in red in the figure) have sensitivity values of
a different distribution than the rest.
When cell line responses on sensitive drugs cannot be accurately
predicted by regression models, it will further lead to imprecise drug selection or prioritization
(e.g., sensitive drugs may be predicted as insensitive). 

Instead, ranking methods represent a more natural and effective alternative so as
to directly prioritize and select drugs.
In order to enable drug selection, in the end, a sorted/ranking order of
drugs needs to be in place.
Accurate predicted cell line response values on drugs can serve to
sort/rank drugs in order. However, any other scores can also serve the purpose as long as they 
produce desired drug orders. This provides the opportunity for learning-to-rank
methods for drug selection, which focus on learning the drug ranking structures directly
(via using certain scores to sort drugs).
Actually, regression based drug selection corresponds to point-wise learning to
rank, 
which has been demonstrated~\cite{Cao2007}
to perform suboptimally compared to
pairwise  
and listwise ranking methods.  
Detailed literature review on learning to rank is available in Section~\ref{sec:work}
\hlgreen{(additional references are available in Section{~\ref{supp:sec:refs}}
  in the supplementary materials)}. 
%

%
The rest of the manuscript is organized as follows.
Section~\ref{sec:work} presents the literature review on learning-to-rank methods.
Section~\ref{sec:methods} presents the new \MTCRP method.
Section~\ref{sec:materials} presents the materials used in experiments.
Section~\ref{sec:experiments} presents the experimental results.
Section~\ref{sec:conclusion} presents the conclusions.

\section{Literature Review on Learning to Rank}
\label{sec:work}
Learning to Rank (\LETOR)~\cite{Furnkranz2010} 
focuses on developing machine learning methods and models that can produce accurate rankings of
interested instances,
rather than using pre-defined scoring functions to sort the instances. 
\LETOR is the key enabling 
technique in information retrieval~\cite{Li2011}.
Existing \LETOR methods fall into three categories:  
1). pointwise methods~\cite{Cao2007}, which learn a score on each individual instance that
will be used to sort/rank all the instances; 
2). pairwise methods~\cite{Burges2007}, which optimize
pairwise ranking orders among all instances to induce good ranking orders among
the instances; and
3) listwise methods~\cite{Lebanon2002}, which model the full
combinatorial structures of ranking lists.
It has been demonstrated~\cite{Cao2007} that 
pairwise and listwise ranking methods outperform pointwise methods in general.
This is because in pairwise and listwise methods, the ordering structures among
instances are leveraged in learning, whereas in pointwise methods, no ordering information is used.
Moreover, listwise methods are more computationally challenging than the others,
due to the combinatorial nature of ranking lists as a whole.
Thus, pairwise methods are the choice in many ranking problems, given the trade-off
between ranking performance and computational demands.

The idea of using \LETOR approaches to prioritize compounds has also drawn some recent
attention~\cite{Liu2017, Zhang2015, Liu2017a}.
For example, Agarwal \etal~\cite{Agarwal2010} developed bipartite
ranking~\cite{Agarwal2005} to rank chemical structures for Structure-Activity-Relationship (SAR)
modeling such that active compounds and inactive compounds are well separated in the ranking lists.
Liu and Ning~\cite{Liu2017a} developed a ranking method with
bi-directional powered push strategy to prioritize selective compounds from multiple bioassays. 
However, \LETOR has not been widely used in prioritizing drugs in computational medicine
domain. 

In \LETOR, a particular interest is to improve the performance on the top of the
ranking lists~\cite{Boyd2012, Narasimhan2013},
that is, instead of optimizing the entire ranking
structures, only the top of the ranking lists 
will be optimized (i.e., to rank the most relevant instances on top), while
the rest of the ranking lists, particularly the bottom of the ranking lists, is of
little interest. 
An effective technique to enable good ranking performance on top in \LETOR is via
push~\cite{Rudin2009, Agarwal2011, Liu2017}.
%
The key idea is to explicitly push relevant instances onto top during optimization. 
Various optimization algorithms are developed to deal with the non-trivial objective functions 
when push is involved~\cite{Burges2007,Li2014}. 
%

%
\section{Methods}
\label{sec:methods}

%
We propose the joint push and LEarning TO Rank with genomics regularization (\MTCRP)
for drug prioritization and selection.
The \MTCRP method learns and uses latent vectors of drugs and cell lines to
score each drug in a cell line,
and ranks the drugs based on their scores (Section~\ref{sec:methods:prediction}).
During the learning process, \MTCRP explicitly pushes the sensitive drugs on
top of the ranking lists that are produced by the prospective latent vectors
(Section~\ref{sec:methods:ppush}), and optimizes the ranking orders among sensitive
drugs (Section~\ref{sec:methods:order}) simultaneously.
In addition, \MTCRP uses genomics information on cell lines to constrain cell line
latent vectors (Section~\ref{sec:methods:opt}).
The following sections describe \MTCRP in detail. The supplementary materials
are available online\footnote{http://cs.iupui.edu/$\sim$liujunf/projects/CCLERank/}. 

%
\begin{table}[!h]
  \begin{center}
  \begin{threeparttable}
  \caption{\mbox{Notations}}
  \label{table:notations}
  \begin{small}
      \begin{tabular}{
            @{\hspace{10pt}}l@{\hspace{10pt}}
            @{\hspace{10pt}}l@{\hspace{10pt}}
      }
        \toprule
          notation & meaning \\
          \midrule
          $\cellline_p$ & cell line $p$ \\
          $\drug_i$     & drug $i$ \\
          $\drug^{\plus}$/$\drug^{\minus}$ & a sensitive/insensitive drug in a cell line\\
          $\cellline_p^{\plus}$/$\cellline_p^{\minus}$ & the set of sensitive/insensitive drugs in $\cellline_p$ \\
          $n_p^{\plus}$/$n_p^{\minus}$ & the size of $\cellline^{\plus}_p$/$\cellline^{\minus}_p$ \\
          $\mathbf{u}_p$/$\mathbf{v}_i$ & latent vector for cell line $\cellline_p$/drug $\drug_i$\\
          $m$/$n$ & the total number of cell lines/drugs \\
          \bottomrule
      \end{tabular}
  \end{small}
  \end{threeparttable}
  \end{center}
\vspace{-10pt}
\end{table}

Table~\ref{table:notations} presents the key notations used in the manuscript. 
In this manuscript, 
drugs are indexed by $i$ and $j$, and cell lines are indexed by $p$ and $q$.
We use $\drug^{\plus}$/$\drug^{\minus}$ to indicate sensitive/insensitive drugs
(sensitivity labeling will be discussed later in Section~\ref{sec:materials:dataset:labelling})
in a certain cell line, for example, 
$\drug_i^{\plus} \in \cellline_p$ or $\drug_i \in \cellline^{\plus}_p$
indicates that drug $\drug_i$ is sensitive in cell line
$\cellline_p$. Cell line is neglected when no ambiguity arises. 
%

%
%
%

\subsection{Drug Scoring}
\label{sec:methods:prediction}

We model that the ranking of drugs in terms of their sensitivities in a cell line
is determined by their latent scores in the cell line.
The latent score of drug $\drug_i$ in cell line $\cellline_p$, denoted as $f_p(\drug_i)$,
is estimated as the dot product of $\drug_i$'s latent vector
$\mathbf{v}_i \in \mathbb{R}^{l\times 1}$
and $\cellline_p$'s latent vector $\mathbf{u}_p \in \mathbb{R}^{l\times 1}$,
where $l$ is the latent dimension, that is, 
\begin{eqnarray}
  \label{eqn:pred}
  \begin{aligned}
    f_p(\drug_i) = f(\drug_i, \cellline_p) = \mathbf{u}^{\mathsf{T}}_p \mathbf{v}_i, 
  \end{aligned}
\end{eqnarray}
where $f(\drug,\cellline)$ is the dot-product scoring function, and 
the latent vectors $\mathbf{u}_p$ and $\mathbf{v}_i$ will be learned.
Then all the drugs are sorted based on their scores in $\cellline_p$.
%
The most sensitive drugs in a cell line will have the highest scores and 
will be ranked higher than insensitive drugs. 
Thus, drug selection in \MTCRP is to identify optimal drug and cell line
latent vectors that together produce preferable cell line-specific drug scores and rankings. 
Note that in \MTCRP, we look for scores $f_p(\drug_i)$
as long as they can produce correct drug rankings, but these scores are not 
necessarily identical to drug sensitivity values (e.g., shifted drug sensitivity
values can also produce perfect drug rankings).
\hlgreen{In addition, the scoring scheme is similar to that used in some matrix factorization approaches
  in recommender systems research{~\cite{Shi2014}}. However, the latent vectors
  $\mathbf{u}$ and $\mathbf{v}$ will be learned via learning the top ranking structures (as will
  be discussed later), not through factorizing the cell line-drug matrix. }

\subsection{Pushing up Sensitive Drugs}
\label{sec:methods:ppush}

%
To enforce the high rank of sensitive drugs, we leverage the idea of ranking with
push~\cite{Agarwal2011}.
The key idea is to quantitatively measure the ranking positions of
drugs, and look for ranking models that can optimize such quantitative
measurement so as to rank sensitive drugs high and insensitive drugs low.
In \MTCRP, we use the height of an insensitive drug $\drug^{\minus}_i$ in $\cellline_p$, 
denoted as $h_f(\drug_i^{\minus}, \cellline_p)$,
to measure its ranking position in $\cellline_p$~\cite{Agarwal2011} as follows, 
\begin{equation}
  \label{eqn:height}
  h_f(\drug_i^{\minus}, \cellline_p) = \sum_{\scriptsize{\drug_j^{\plus}\in \cellline_p^{\plus}}}
  \mathbb{I} ( f_p(\drug_j^{\plus}) \le f_p(\drug_i^{\minus}) ),  
\end{equation}
where $\cellline_p^{\plus}$ is the set of sensitive drugs in cell line $\cellline_p$,
$f$ is the drug scoring function (Equation~\ref{eqn:pred}),
$f_p(\drug_j^{\plus})$ and $f_p(\drug_i^{\minus})$ are the scores of $\drug_j^{\plus}$ and $\drug_i^{\minus}$ in
$\cellline_p$, respectively, 
and
$\mathbb{I}(x)$ is the indicator function 
($\mathbb{I}(x) = 1$ if $x$ is true, otherwise 0).
%
Essentially, $h_f(\drug_i^{\minus}, \cellline_p)$
is the number of sensitive drugs that are ranked below the
insensitive drug $\drug^{\minus}_i$ in cell line $\cellline_p$ by the scoring function $f$. 

To push sensitive drugs higher in a cell line, it is to minimize the total
height of all insensitive drugs in that cell line
\hl{
To push sensitive drugs higher in a cell line, the total
height of all insensitive drugs in that cell line should be minimized
}
(i.e., minimize the total number of sensitive drugs that are ranked below 
insensitive drugs). 
%
%
Thus, for all the cell lines, it is to minimize their total heights, denoted as
$P^{\uparrow}_f$, that is, 
\vspace{-10pt}
\begin{equation}
  \label{eqn:push}
  P^{\uparrow}_f = \sum_{p = 1}^m
   \frac{1}{n_p^{\plus}n_p^{\minus}}\sum_{\scriptsize{\drug^{\minus}_i\in \cellline_p}}h_f(\drug_i^{\minus}, \cellline_p), 
\end{equation}
where $m$ is the number of cell lines,
and $n^{\plus}_p$ and $n^{\minus}_p$ are the numbers of sensitive and insensitive drugs in cell
line $\cellline_p$.
%
The normalization by $n^{\plus}_p$ and $n^{\minus}_p$ is to eliminate the effects from
different cell line sizes.
%

\subsection{Ranking among Sensitive Drugs}
\label{sec:methods:order}

In addition to pushing sensitive drugs on top of insensitive drugs, 
we also consider the ranking orders among sensitive drugs in order to enable fine-grained
prioritization among sensitive drugs. 
Specifically, we use $\drug_i \succ_{\scriptsize{R}} \drug_j$ to represent
that $\drug_i$ is ranked higher than $\drug_j$ in the relation $R$. 
We use concordance index (\CI) 
to measure drug
ranking structures compared to the ground truth, which is defined as follows, 
%
\begin{equation}
  \label{eqn:ci}
  \CI(\{\drug_i\}, \cellline, f) = \frac{1}{|\{\drug_i \succ_{\scriptsize{\cellline}} \drug_j\}|}
  \sum_{\scriptsize{\drug_i \succ_{\tiny{\cellline}} \drug_j}}\mathbb{I}(\drug_i \succ_f \drug_j), 
\end{equation}
%
where $\{\drug_i\}$ is the set of drugs in cell line \cellline,
$\{\drug_i \succ_{\scriptsize{\cellline}} \drug_j\}$
is the set of ordered pairs of drugs in cell line \cellline
($\drug_i \succ_{\scriptsize{\cellline}} \drug_j$ represents that $\drug_i$ is 
more sensitive, and thus ranked higher, than $\drug_j$ in \cellline), 
$f$ is the scoring function (Equation~\ref{eqn:pred}) that produces an estimated drug ranking, 
$\drug_i \succ_f \drug_j$ represents that 
$\drug_i$ is ranked higher than $\drug_j$ by $f$, 
and $\mathbb{I}$ is the indicator function. 
Essentially, \CI measures the ratio of correctly ordered drug pairs by $f$
among all possible pairs.
Higher \CI values indicate better ranking structures. 
To promote correct ranking orders among sensitive drugs in all the cell lines, we minimize the
objective $O^{\plus}_f$, defined as the sum of $1-\CI$ values 
(i.e., the ratio of mis-ordered drug pairs among all pairs) over the sensitive drugs of all 
the cell lines, as follows, 
%
\begin{eqnarray}
  \begin{aligned}
  \label{eqn:order_rel}
%
  O^{\plus}_f  & = \sum_{p=1}^m [1 - \CI(\{\drug_i^{\plus}\}, \cellline_p, f)] \\
     	 & = \sum_{p=1}^m \frac{1}{|\{\drug_i^{\plus} \succ_{\scriptsize{\cellline_p}} \drug_j^{\plus}\}|}
     \sum_{\scriptsize{\drug_i^{\plus} \succ_{\tiny{\cellline_p}} \drug_j^{\plus}}}\mathbb{I}(\drug_i^{\plus} \prec_f \drug_j^{\plus}). 
  \end{aligned}
\end{eqnarray}
\subsection{Overall Optimization Problem for \MTCRP}
\label{sec:methods:opt}

Overall, we seek the cell line latent vectors and drug latent vectors that will be used
in drug scoring function $f$ (Equation~\ref{eqn:pred}) 
such that for each cell line, the sensitive drugs will be ranked on top and in
right orders using the latent vectors.
In \MTCRP, such latent vectors are learned by solving the following optimization problem:
\vspace{-5pt}
\begin{equation}
  \label{eqn:obj}
    \min_{U, V} \mathcal{L}_f = (1 - \alpha) P^{\uparrow}_f + \alpha O^{\plus}_f
  + \frac{\beta}{2} R_{uv} + \frac{\gamma}{2} R_{\text{csim}}, 
\end{equation}
where $\mathcal{L}_f$ is the overall loss function;  
$P^{\uparrow}_f$ and $O^{\plus}_f$ are defined in Equation~\ref{eqn:push} and
Equation~\ref{eqn:order_rel}, respectively; 
\mbox{$U = [\mathbf{u}_1, \mathbf{u}_2, \cdots, \mathbf{u}_m]$}
and \mbox{$V=[\mathbf{v}_1, \mathbf{v}_2, \cdots, \mathbf{v}_n]$}
are the latent vector matrices for cell lines and drugs, respectively
($U \in \mathbb{R}^{l \times m}$, $V \in \mathbb{R}^{l \times n}$, where $l$ is the 
latent dimension);  
$\alpha$ ($\alpha \in [0, 1]$) is a weighting parameter to control the contribution from
push (i.e., $P^{\uparrow}_f$) and ranking (i.e., $O^{\plus}_f$);  
$\beta$ and $\gamma$ are regularization parameters ($\beta \ge 0$,
$\gamma \ge 0$) on the two regularizers $R_{uv}$ and $R_{\text{csim}}$, respectively.

In Problem~\ref{eqn:obj}, $R_{uv}$ is a regularizer on $U$ and $V$ to prevent overfitting,
defined as
\vspace{-5pt}
\begin{equation}
  \label{eqn:reg}
    R_{uv} = \frac{1}{m}\|U\|^2_F + \frac{1}{n}\|V\|^2_F, 
\end{equation}
where $\|X\|_F$ is the Frobenius norm of matrix $X$. 
%
%
%
$R_{\text{csim}}$ is a regularizer on cell lines
to constrain cell line latent vectors, defined as
\begin{equation}
  \label{eqn:sim}
    R_{\text{csim}} = \frac{1}{m^2}\sum_{p=1}^m\sum_{q=1}^m w_{pq}\|\mathbf{u}_p-\mathbf{u}_q\|_2^2,    
\end{equation}
where $w_{pq}$ is the similarity between $\cellline_p$ and $\cellline_q$ that is calculated
using genomics information of the cell lines (e.g., gene expression information).
The underlying assumption is that if two
cell lines have similar patterns in their genomics data (i.e., large $w_{pq}$),
they will be similar in their \hl{cell line response patterns}, 
and thus similar latent vectors~\cite{Costello2014}.
%
%


The Problem~\ref{eqn:obj} involves an indicator function
(in Equation~\ref{eqn:height}, \ref{eqn:ci}), which is not continuous or smooth.
Thus, we use the logistic function as its surrogate~\cite{Rudin2009}, that is, 
%
\begin{equation}
  \label{eqn:surrogation}
  \mathbb{I}(x \le y) \approx \log[1+\exp(-(x-y))]= -\log \sigma(x-y),
\end{equation}
%
%
%
where $\sigma(x)$ is a sigmoid function, that is,
$\sigma(x) = \frac{1}{1+\exp(-x)}$. 
%
%
The optimization algorithm for \MTCRP optimization is presented in Algorithm~\ref{alg:opt} \hl{in supplementary materials}. 
We use alternating minimization with gradient descent (details in
Section~\ref{sec:appendix:gradient_and_update} \hl{in supplementary materials})
to solve the optimization Problem~\ref{eqn:obj}.
%

\begin{figure*}[!h]
  \centering
  \begin{minipage}{0.45\linewidth}
    \begin{center}
      \vspace{-30pt}
      \scalebox{0.9}{
        \input{figures/random_5_fold.tex}
      }
      \vspace{-35pt}
      \definecolor{train}{HTML}{A1A1FF}
\definecolor{test}{HTML}{E2E3FF}

\begin{tikzpicture}[fill=white]

    \draw
    (0.0, 0.0) node {};

    \draw
    [fill=train] (0.6, 0.0) rectangle (1.1, 0.3) 
    (2.6, 0.15) node [text=black] {training drug};

    \draw
    [fill=test] (4.3, 0.0) rectangle (4.8, 0.3)
    (6.3, 0.15) node [text=black] {testing drug};

\end{tikzpicture}
      \vspace{10pt}
      \caption{Data split for 5-fold cross validation}
      \label{fig:5fold}
    \end{center}
  \end{minipage}
  \begin{minipage}{0.45\linewidth}
  \begin{center}
    \vspace{-30pt}
    \scalebox{0.9}{
      \input{figures/random_loo.tex}
    }
    \vspace{-35pt}
      \definecolor{train}{HTML}{A1A1FF}
\definecolor{test}{HTML}{E2E3FF}

\begin{tikzpicture}[fill=white]

    \draw
    (0.0, 0.0) node {};

    \draw
    [fill=train] (0.6, 0.0) rectangle (1.1, 0.3) 
    (2.6, 0.15) node [text=black] {training cell lines};

    \draw
    [fill=test] (4.3, 0.0) rectangle (4.8, 0.3)
    (6.3, 0.15) node [text=black] {testing cell lines};

\end{tikzpicture}
    \vspace{10pt}
    \caption{Data split for testing new cell lines}
    \label{fig:loo}
  \end{center}
  \end{minipage}
\vspace{-10pt}
\end{figure*}

%
%


Since the number of drugs pairs is quadratically larger than the number of drugs, 
it could be computationally
expensive to use all the drug pairs during training. To solve this issue, we develop
a sampling scheme. 
%
During each iteration of training, we use all the sensitive drugs in each cell line but
randomly sample a same number of insensitive drugs from each respective cell line.
This process is repeated for a number of times
and then the average gradient is used to update $U$ and $V$.
This sampling scheme will significantly speed up the optimization process. 

%

\section{Materials}
\label{sec:materials}

\subsection{Dataset and Experimental Protocol}
\label{sec:materials:dataset}

%
%
\begin{table}[h!]
  \centering
  \caption{Dataset Description}
  \begin{threeparttable}
      \begin{small} 
      \begin{tabular}{
          @{\hspace{6pt}}c@{\hspace{6pt}}
          @{\hspace{6pt}}c@{\hspace{6pt}}
          @{\hspace{6pt}}c@{\hspace{6pt}}
          @{\hspace{6pt}}c@{\hspace{6pt}}
          @{\hspace{6pt}}c@{\hspace{6pt}}
          @{\hspace{6pt}}c@{\hspace{6pt}}
          @{\hspace{6pt}}c@{\hspace{6pt}}
        }
        \toprule
        m & n & \#genes & \#AUCs & \#mAUCs & \#\drug/\cellline & \#\cellline/\drug \\
        \midrule
        821 & 545 & 20,068 & 357,052 & 90,393 & 435 & 655 \\
        \bottomrule
      \end{tabular}
      \begin{tablenotes}
        \setlength\labelsep{0pt}
        \begin{scriptsize}
        \item
          The columns of ``m'', ``n'' and ``\#genes'' have the number of
          cell lines, drugs and genes in the dataset, respectively.
          The columns of ``\#AUCs'' and ``\#mAUCs'' have the total number of available
          response values and missing response values, respectively.
          The column of ``\#\drug/\cellline'' has the average number of available drug
          response values per cell line.
          The column of ``\#\cellline/\drug'' has the average number of cell lines that have
          response values for each drug. 
          \par
        \end{scriptsize}
      \end{tablenotes}
      \end{small}
  \end{threeparttable}
  \label{table:dataset}
\vspace{-10pt}
\end{table}

%
We use the cell line data and drug sensitivity data from
Cancer Cell Line Encyclopedia (CCLE)~\footnote{https://portals.broadinstitute.org/ccle/home}
and Cancer Therapeutics Response Portal (CTRP v2)~\footnote{https://portals.broadinstitute.org/ctrp/}
(both accessed on 10/14/2016), respectively.
%
%
CTRP provides the cell line responses to different drugs. The response
is measured using area-under-concentration-response curve (AUC) sensitivity
scores. 
Lower response (AUC) scores indicate higher drug sensitivities. 
CCLE provides the expression information over a set of genes for each of the cell lines.
Larger expression values indicate higher gene expression levels.
CCLE also provides other omics data for the cell lines (e.g., copy number variations).
In this manuscript, we only use gene expression information, as it is demonstrated as the
most pertinent to cell line response~\cite{Costello2014}.
The use of other omics data will be explored in the future research.
This dataset has large numbers of both cell lines and drugs. 
Table~\ref{table:dataset} presents the description of the dataset used in the experiments.
Note that in the dataset, about 20\% of the drug sensitivity values are missing. For the
drugs which do not have response values in a cell line, we do not use the drugs in
learning the corresponding cell line latent vector.

%
\subsubsection{Experimental Setting}
\label{sec:materials:dataset:split}

We had two experimental settings for the experiments.

%
\paragraph{N-Fold Cross Validation}
\label{sec:materials:dataset:split:5fcv}
%
%
%
%
In the first setting, 
we split drug sensitivity data for each cell line into a training and a testing set, 
and conduct 5-fold cross validation to evaluate model performance.
Fig.~\ref{fig:5fold} demonstrates the training-testing splits. 
For each cell line, its drug sensitivity data are randomly split into 5 folds.
One of the 5 folds is used as testing set and the other four folds are used for training.
\hl{This is done 5 times, until each of the five folds has served as testing data.}
The final results are the average over the 5 folds.
This experimental setting corresponds to the application scenario in which
additional drugs (i.e., the testing data) need to be selected for each cell line/patient. 
%

%
During the data split, we ensure that for each of the drugs, there is at least one
cell line in the training set that has response information for that drug. This is to
avoid the situation in which drugs in the testing set do not have information
during training, or the use scenario in which brand-new compounds need to be selected for further
testing. The latter will be studied \hl{in future research}.
We also ensure that each cell line has drug sensitivities \hl{in the training set} to avoid the 
situation of brand-new cell lines. 
This situation will be studied in the second experimental setting. 
%

%
%
\paragraph{Leave-Out Validation}
\label{sec:materials:dataset:split:locv}
%
%
%
%
We also conduct experiments in a different setting as indicated in Fig.~\ref{fig:loo},
that is, we hold out entire cell lines into the testing data so that in training data, the
held-out cell lines have no drug response information at all.
This corresponds to the use scenario to select sensitive drugs for new cell line/new patients. 
Details on how to hold out cell lines will be discussed later in
Section~\ref{sec:experiments:holdout}.

\subsubsection{Sensitivity Labeling Scheme}
\label{sec:materials:dataset:labelling}

\paragraph{Labeling Scheme for N-Fold Cross Validation}
\label{sec:materials:dataset:labelling:5fold}

In the 5-fold setting (Fig.~\ref{fig:5fold}),
for each cell line, we use a certain percentile $\theta$ (e.g., \mbox{$\theta$=5})
of all its response values in the training set as a threshold to determine drug
sensitivity in that cell line.
Thus, the sensitivity threshold is cell line specific. It is only selected from 
the training data of respective cell lines
(i.e., testing data are not used to determine the threshold as they are considered
as unknown during training). 
Drugs in both the training set and testing set are then labeled as sensitive in
the respective cell line if the cell line has lower response values 
on the drugs than the threshold (lower AUC scores indicate higher sensitivity),
otherwise, the drugs are labeled as insensitive. 
The reason why a cell-line-specific percentile threshold is used for sensitivity
labeling is that there lacks a pre-defined threshold of sensitivity scores for
each of the cell lines to determine sensitivity labels.
Meanwhile, given the heterogeneity of cell lines, we cannot \hl{apply the same} threshold
for different cell lines. 
The idea of using sensitivity score percentile as a threshold is very similar to
that in Speyer \etal~\cite{Speyer2017}, in which the outliers with low sensitivity
scores are labeled as sensitive. 

%
\paragraph{Labeling Scheme for Testing New Cell Lines}
\label{sec:materials:dataset:labelling:new}

In the second setting with new cell lines (Fig.~\ref{fig:loo}), since the new cell lines
have no drug response information in training, we use a percentile threshold from the
testing data (i.e., the new cell lines; the ground truth) to label sensitivities
of the drugs in the new cell lines.

\subsection{Baseline Methods}
\label{sec:materials:baseline}
%
\hl{We use two strong baseline methods for comparison: 
the Bayesian Multi-Task Multi-Kernel Learning ({\BMTMKL}) method
and the Kernelized Rank Learning ({\KRL}) method}.

\subsubsection{Bayesian Multi-Task Multi-Kernel Learning (\BMTMKL)}
\label{sec:materials:baseline:bmtmkl}

%
%
\BMTMKL{~\cite{Costello2014}}, winning method for DREAM 7
challenge~\footnote{http://dreamchallenges.org/project/dream-7-nci-dream-drug-sensitivity-prediction-challenge/}, was originally developed to rank cell lines with respect to a drug based on
their responses to the drug. 
In \BMTMKL, cell line ranking for each drug is considered a task.
All the cell line rankings are learned simultaneously in a multi-task
learning 
framework.
Multiple kernels 
are constructed from multiple types of omics data
for cell lines to quantify their similarities.  
The multi-task and multi-kernel learning is conducted within
a kernelized regression with Bayesian inference for parameter estimation. 
Thus, \BMTMKL predicts drug response values via the regression
and uses the values for cell line ranking, that is,
\BMTMKL is a point-wise learning-to-rank method. 

Note that the drug ranking problem we are tackling in this manuscript
is a different problem compared to the cell line ranking
problem that \BMTMKL is designed to tackle.
The cell line ranking problem 
corresponds to the application scenario in which 
cell lines/patients need to be selected to test a given drug. 
The drug ranking problem corresponds to the application scenario in which 
drugs need to be selected to treat a given cell line/patient. 
%
\hl{Still, {\BMTMKL} can be used on drug ranking problems}
by switching the roles of ``drugs'' and ``cell lines'', and 
the predicted drug response values can also be used for drug ranking.
%

\subsubsection{Kernelized Rank Learning (\KRL)}
\label{sec:materials:baseline:krl}

\hlgreen{%
  {\KRL}{~\cite{He2018}} learns a ranking structure of drugs in each cell line by approximating
  the sensitivity values of drugs in the cell line via a kernelized regression. 
  The kernel matrix is constructed from the gene expression profiles of
  cell lines. 
  During learning, {\KRL} optimizes
  the Normalized Discounted Cumulative Gain (NDCG@$k$){~\cite{Liu2009}}, a
  popular metric in learning to rank, 
  to rank the most sensitive drugs of each cell line among top $k$ of the drug ranking list. 
  Therefore, {\KRL} is a list-wise learning-to-rank method. 
}

\subsection{Evaluation Metrics}
\label{sec:materials:evaluation}

%
%
We first introduce the evaluation metrics that are used in most of the experiments.
Other metrics that are used in specific experiments will be introduced later when they
are applied. 
The first metric that we use to evaluate the performance of \BMTMKL and \MTCRP
is the average-precision at $k$ (AP@$k$)~\cite{Liu2009}.
It is defined as the average of precisions that are calculated at each ranking position
of sensitive drugs that are ranked among top $k$
in a ranking list, that is, 
%
\begin{equation}
  \label{eqn:ap@k}
  \text{AP}@k(\{\drug_i\}, \cellline, f) =
  \small{
  \frac{
    \displaystyle\sum_{j=1}^k
        \text{Prec}(\{\drug_{\overrightarrow{1}}, \cdots, \drug_{\overrightarrow{j}}\}, \cellline^{\plus}, f) \cdot
    \mathbb{I}(\drug_{\overrightarrow{j}} \in \cellline^{\plus})
  }
  {
    \displaystyle\sum_{j=1}^k
    \mathbb{I}(\drug_{\overrightarrow{j}} \in \cellline^{\plus})
  }
  }
  ,
\end{equation}
where $\drug_{\overrightarrow{j}}$ is the drug that is ranked at position $j$ by $f$,
$\mathbb{I}(\drug_{\overrightarrow{j}} \in \cellline^{\plus})$
checks whether $\drug_{\overrightarrow{j}}$ is sensitive in \cellline in the
ground truth, and 
Prec is defined as
\begin{equation}
  \text{Prec}(\{\drug_{\overset{\scalebox{0.6}{$\rightharpoonup$}}1}, \cdots, \drug_{\overrightarrow{j}}\}, \cellline^{\plus}, f) = \sum_{i=1}^j
  \mathbb{I}(\drug_{\overset{\scalebox{0.6}{$\rightharpoonup$}}i} \in \cellline^{\plus})\big/ j, 
\end{equation}
that is, it is calculated as the ratio of sensitive drugs among top-$j$ ranked drugs. 
Thus, AP@$k$ considers the ranking positions of
sensitive drugs that are ranked among top $k$ of a ranking list. It is
a popular metric to evaluate \LETOR methods. 
Higher AP@$k$ values indicate that the sensitive drugs are ranked higher on average.

We define a second metric average-hit at $k$ (AH@$k$) as the average number of sensitive
drugs that are ranked among top $k$ of a ranking list, that is,
\begin{equation}
  \label{eqn:ah@k}
  \text{AH}@k(\{\drug_i\}, \cellline, f) = \sum_{j=1}^k \mathbb{I}(\drug_{\overrightarrow{j}} \in \cellline^{\plus}).
\end{equation}
Higher AH@$k$ values indicate that more sensitive drugs are ranked among top $k$.

We also use \CI as defined in Equation~\ref{eqn:ci} to evaluate the ranking structures
among only sensitive drugs. In this case, we denote \CI specifically as \rCI (i.e.,
\CI for sensitive drugs),
and thus by default, \CI evaluates the entire ranking structures of both sensitive
and insensitive drugs, and \rCI is only for sensitive drugs. 
Note that \rCI (\CI) and AP@$k$ measure different aspects of a
ranking list. The \rCI (\CI) metric measures whether the ordering structure of a
ranking list is close to its ground truth,
while AP@$k$ measures whether the relevant instances
(i.e., sensitive drugs in this manuscript) \hl{are} ranked on top.
A high AP@$k$ does not necessarily indicate the ordering among the top-ranked
drugs is correct. Similarly, a high \rCI (\CI) does not necessarily lead to
that the most sensitive drugs \hl{being} ranked on top, particularly when there are many
insensitive drugs in the list. 
In this manuscript, both the drug sensitivity and the ordering of
sensitive \hl{drugs are of concern}. 
%
%
\hl{
That is, we would like to ensure that sensitive drugs are ranked 
higher than insensitive drugs, and meanwhile, the sensitive drugs are well
ordered in terms of their relative sensitivities (i.e., among all sensitive drugs,
drugs that are more sensitive
are ranked higher than those that are less sensitive). 
}

%


\begin{table*}[t!]
  \centering
  \caption{\mbox{Performance on Ranking New Drugs}}
  {
  \label{tbl:results:overall_2p5p}
  \begin{small}
  \begin{threeparttable}
      \begin{tabular}{
        @{\hspace{8pt}}c@{\hspace{8pt}}
        @{\hspace{8pt}}c@{\hspace{8pt}}
        @{\hspace{0pt}}r@{\hspace{0pt}}
        @{\hspace{8pt}}r@{\hspace{8pt}}
        @{\hspace{8pt}}r@{\hspace{8pt}}
        @{\hspace{8pt}}r@{\hspace{8pt}}
        @{\hspace{8pt}}c@{\hspace{8pt}}
        @{\hspace{8pt}}r@{\hspace{8pt}}
        @{\hspace{8pt}}r@{\hspace{8pt}}
        @{\hspace{8pt}}r@{\hspace{8pt}}
        @{\hspace{8pt}}r@{\hspace{8pt}}
        @{\hspace{8pt}}r@{\hspace{8pt}}
        @{\hspace{8pt}}r@{\hspace{8pt}}
        @{\hspace{8pt}}r@{\hspace{8pt}}          
        }        
        \toprule
        sthr	& method   & \multicolumn{6}{c}{parameters} & \multicolumn{6}{c}{performance} \\
        \midrule
	\multirow{10}{*}{$\theta = 2$}

        & \multirow{2}{*}{\BMTMKL}
        & \multicolumn{2}{c}{$\alpha_b$}  & \multicolumn{2}{c}{$\beta_b$} & usim & $\sigma$       & \small{AP@$5$} & \small{AH@$5$}	& \small{AP@$10$} & \small{AH@$10$}	& \small{\rCI}   & \small{\CI}               \\

        \cmidrule(r){3-14}                                                                                                                                                             
        & & \multicolumn{2}{c}{1.0e-10} & \multicolumn{2}{c}{1.0e+10} 	& \hl{$\RBF$} 	& 10.0 	  & \textbf{0.740} &  \textbf{1.702}	&  \textbf{0.711}  & \textbf{2.072} & \textbf{0.646}	& \textbf{\underline{0.812}}\\

        \cmidrule(r){2-14}

        & \multirow{3}{*}{\hlgreen{{\KRL}}}
        & \multicolumn{2}{c}{$k$}  & \multicolumn{2}{c}{$\lambda$} & usim & $\sigma$       & \small{AP@$5$} & \small{AH@$5$}     & \small{AP@$10$} & \small{AH@$10$}     & \small{\rCI}   & \small{\CI}               \\

        \cmidrule(r){3-14}                                                                                                                                             
        & & \multicolumn{2}{c}{10} & \multicolumn{2}{c}{1.0e-06}   & \hl{$\RBF$}   & 0.001    & \textbf{0.753} &  \textbf{1.784}    &  \textbf{0.725}  & \textbf{2.137} & 0.673                         & 0.702\\

        & & \multicolumn{2}{c}{10} & \multicolumn{2}{c}{1.0e-05}   & $\LIN$        & -        & 0.668          &  1.620             &  0.642           & 2.065          & \textbf{\underline{0.683}}    & \textbf{0.745}\\

        \cmidrule(r){2-14}

        & \multirow{5}{*}{\MTCRP}
        & $l$  & $\alpha$ & $\beta$ & $\gamma$ & usim & $\sigma$ & \small{AP@$5$}             & \small{AH@$5$}              & \small{AP@$10$}                  & \small{AH@$10$}               & \small{\rCI}            & \small{\CI}          \\
        \cmidrule(r){3-14}                                                                                                                                                                                                                                 
        & & 50 & 1.0      & 1.0     & 100.0    & \hl{$\COS$}  & - 	 & 0.686                      & 1.606                       & 0.663 	                       & 1.938                         & \textbf{0.680} & 0.770                \\
        & & 30 & 0.1      & 1.0     & 100.0    & \hl{$\RBF$}  & 10.0     & 0.527                      & 1.291                       & 0.505                            & 1.809                         & 0.505                      & \textbf{0.805}       \\
        & & 10 & 0.0      & 0.1     & 100.0    & \hl{$\COS$}  & -        & \underline{\textbf{0.783}} & \underline{\textbf{1.856}}  & \underline{\textbf{0.758}}       & \underline{\textbf{2.159}}    & 0.639                      & 0.774                \\

        & & 10 & 0.0      & 0.1     &   \hl{0.0}    & {$\COS$}  & -  & 0.780 & 1.851 & 0.755 & 2.157 & 0.631 & 0.786                \\

        \midrule

        \multirow{12}{*}{$\theta = 5$}

        & \multirow{2}{*}{\BMTMKL}
        & \multicolumn{2}{c}{$\alpha_b$}  & \multicolumn{2}{c}{$\beta_b$} & usim & $\sigma$      & \small{AP@$5$} & \small{AH@$5$}     & \small{AP@$10$}  & \small{AH@$10$} & \small{\rCI}   & \small{\CI}               \\
        \cmidrule(r){3-14}                                                                                                                                                             
        %
        & & \multicolumn{2}{c}{1.0e-10} & \multicolumn{2}{c}{1.0e+10}   & \hl{$\RBF$}   & 10.0           & \textbf{0.828} &  \textbf{2.736}    &  \textbf{0.772}  & \textbf{3.761} & \textbf{0.652} & \textbf{\underline{0.812}}\\

        \cmidrule(r){2-14}

        & \multirow{4}{*}{\hlgreen{{\KRL}}}
        & \multicolumn{2}{c}{$k$}  & \multicolumn{2}{c}{$\lambda$} & usim & $\sigma$       & \small{AP@$5$} & \small{AH@$5$}     & \small{AP@$10$} & \small{AH@$10$}     & \small{\rCI}   & \small{\CI}               \\

        \cmidrule(r){3-14}                                                                                                                                             
        & & \multicolumn{2}{c}{10} & \multicolumn{2}{c}{1.0e-06}  & \hl{$\RBF$}   & 0.001    & \textbf{0.817} &  \textbf{2.715}    &  \textbf{0.761}  & \textbf{3.798} & 0.676                         & 0.762\\
        & & \multicolumn{2}{c}{10} & \multicolumn{2}{c}{1.0e-04}  & \LIN          & -        & 0.789          &  2.580             &  0.730           & 3.690          & \textbf{0.689}                & 0.756\\
        & & \multicolumn{2}{c}{10} & \multicolumn{2}{c}{1.0e-05}  & \hl{$\RBF$}   & 0.0001   & 0.796          &  2.547             &          0.736   &         3.560  &                    0.660      & \textbf{0.768}\\

        \cmidrule(r){2-14}

        & \multirow{6}{*}{\MTCRP}
          & $l$  & $\alpha$ & $\beta$ & $\gamma$ & usim & $\sigma$ & \small{AP@$5$}             & \small{AH@$5$}             & \small{AP@$10$}            & \small{AH@$10$}            & \small{\rCI}                   & \small{\CI}                      \\
        \cmidrule(r){3-14}                                                                                                                                                                                                    
        & & 50 & 1.0      &    1.0  & 100.0    & \hl{$\RBF$}  &   10.0   & 0.780  	              & 2.376                      & 0.721                      & 3.228                      & \underline{\textbf{0.699}}     &  0.726                           \\ 
        & & 30 & 0.5      &    0.1  & 100.0    & \hl{$\COS$}  &   -     & 0.744                      & 2.461                      & 0.687                      & 3.581                      & 0.516                          &  \textbf{0.810}                  \\ 
        & & 50 & 0.0      &    0.1  & 100.0    & \hl{$\COS$}  &   -     & \underline{\textbf{0.857}} & 2.919                      & 0.805                      & 3.934                      & 0.663                          &  0.742                           \\ 
        & & 10 & 0.5      &    1.0  & 100.0    & \hl{$\RBF$}  &   10.0   & 0.855 & \underline{\textbf{2.965}} & \underline{\textbf{0.806}} & \underline{\textbf{3.986}}                      & 0.658                          &  0.804                           \\ 

        & & 10 & 0.5      & 0.1     &   \hl{0.0}    & {$\COS$}  & -  & 0.855 & 2.965 & 0.806 & 3.985 & 0.658 & 0.804                \\

        \bottomrule
      \end{tabular}
      \begin{tablenotes}
        \begin{scriptsize}
          \setlength\labelsep{0pt}
        \item 
		%
		%
		The columns corresponding to ``$\alpha_b$'', ``$\beta_b$'',
                ``usim'', and ``$\sigma$'' have the  
		two hyperparameters, cell line similarity function,
                and parameter for \hl{$\RBF$} cell line similarity, respectively, for \BMTMKL.
                \hlgreen{The columns corresponding to ``$k$'', ``$\lambda$'',
                ``usim'', and ``$\sigma$'' have the
                two hyperparameters, cell line similarity function,
                and parameter for {$\RBF$} cell line similarity, respectively, for {\KRL}.}
		The columns corresponding to ``$l$'', ``$\alpha$'',
                ``$\beta$'', ``$\gamma$'', ``usim'', and ``$\sigma$'' have the
                latent dimension, weighting factor,
                latent vector regularization parameter,
                cell line similarity regularization parameter,
                cell line similarity function, and parameter for \hl{$\RBF$} cell line similarity,
                respectively, for \MTCRP. 
		%
		The best performance of each method under each metric is in \textbf{bold}.
		The best performance of both the methods under each metric is \underline{underscored}.     
          \par
        \end{scriptsize}
      \end{tablenotes}
  \end{threeparttable}
  \end{small}
  } 
\vspace{-10pt}
\end{table*}


\subsection{Gene Selection and Cell Line Similarities}
\label{sec:materials:feature}

We use gene expression information to measure cell line similarities (i.e., $w_{pq}$
as in Equation~\ref{eqn:sim}) and
regularize our ranking models (i.e., $w_{pq}\|\mathbf{u}_p-\mathbf{u}_q\|_2^2$
as in Equation~\ref{eqn:sim}).
%
It is well accepted that not all the genes are informative to cell line response to
drugs~\cite{Costello2014}, and thus 
we use $\ell_1$ regularized linear regression to conduct 
feature selection over gene expression data to select informative genes with respect
to each drug. 
It is well known that the $\ell_1$ regularization will promote sparsity in the
solution, 
in which the non-zero values will indicate useful
independent variables (in our case, genes).
To select informative genes, the gene expression values over all the cell lines are considered
as independent variables and the response values on each drug from all the cell lines
are considered as dependent variables.
%
If a cell line has no response value on a drug, the gene expression information of that
cell line is not used. 
A linear least-squares
regression with $\ell_1$ and $\ell_2$ regularization (i.e., elastic net) is applied over these
variables so as to select informative genes for each drug. 
The regularization parameters over the $\ell_1$ regularizer and
the $\ell_2$ regularizer are identified via regularization path~\cite{Park2007}.
Fig.~\ref{supp:fig:gene-selection} in the supplementary materials
demonstrates the regression method for gene selection. 
The union of all the selected genes for all the drugs will be used to calculate cell line
similarities. 
In the end, 1,203 genes are selected. The list of the selected genes is available
Section~\ref{supp:sec:gene} in the supplementary materials.
%
%
We use cosine similarity function (\hl{$\COS$}) and radial basis function (\hl{$\RBF$})
over the selected genes (these genes are considered as cell line features) to calculate
the similarities between cell lines.
\hlgreen{For \KRL, we also use a linear kernel ({\LIN}), following the experimental setting as in
He {\etal}{~\cite{He2018}}, to calculate cell line similarities over
the selected genes. }

\section{Experimental Results}
\label{sec:experiments}

\subsection{Ranking New Drugs}
\label{sec:experiments:overall}

%
We first compare the performance of \BMTMKL, \hl{\KRL} and \MTCRP on
ranking new drugs in each cell line (i.e., ranking testing drugs among themselves
in each cell line). 
The experiments follow the protocol as indicated in Fig.~\ref{fig:5fold}. 
Note that notion of ``new drugs'' is with respect to each
cell line, and a new drug in a cell line could be known in a different cell line.

We use 2 percentile (i.e., $\theta$=2) and 5 percentile (i.e., $\theta$=5) as discussed
in Section~\ref{sec:materials:dataset:labelling} to label sensitivity. 
\hl{Although the baseline methods and {\MTCRP}}
do not rely on specific labeling schemes, the small percentiles make the drug selection
problem realistic. This is because in real practice, only the top few most sensitive drugs
will be of great interest. However, given that the sensitive drugs are few, the drug
selection problem is very non-trivial. 

%
\hl{For the baseline methods and {\MTCRP}}
we conduct a grid search for each of their parameters, and
present the results that correspond to the best parameter combinations.
\hlgreen{The full set of experimental results over all parameters is available in
Table{~\ref{supp:tbl:overall:2percent:baseline}},
{\ref{supp:tbl:overall:5percent:baseline}},
{\ref{supp:tbl:overall:2percent:KRL}},
{\ref{supp:tbl:overall:5percent:KRL}},
{\ref{supp:tbl:overall:2percent}} and 
{\ref{supp:tbl:overall:5percent}}
in the supplementary materials.}
%
Table~\ref{tbl:results:overall_2p5p} presents the overall performance. 

\begin{figure*}[!h]
  \centering
  \begin{minipage}{\linewidth}
    \begin{center}
      \scalebox{0.9}{\input{plot/plot_2prc_legend.tex}}
    \end{center}
    \vspace{-20pt}
  \end{minipage}
  \mbox{
  \begin{minipage}{0.5\linewidth}
    \mbox{
      \begin{minipage}{0.49\linewidth}
        \centering
        \subfigure[\small{$\theta$=2, $l$=10}]{
          {\input{plot/d_10_2prc_alpha.tex}}
          \label{fig:plot-2prc-alpha-ap-change}
        }
      \end{minipage}
      \begin{minipage}{0.49\linewidth}
        \centering
        \subfigure[\small{$\theta$=5, $l$=10}]{
          {\input{plot/d_10_5prc_alpha.tex}}
          \label{fig:plot-5prc-alpha-ap-change}
        }
      \end{minipage}
    }
    \caption{Performance of \MTCRP w.r.t. the Push Parameter $\alpha$}
    \label{fig:alpha}
  \end{minipage}
  \begin{minipage}{0.5\linewidth}
    \mbox{
      \begin{minipage}{0.49\linewidth}
        \centering
        \subfigure[\small{$\theta$=2, $\alpha$=0.0}]{
          {\input{plot/eval_d_2prc.tex}}
          \label{fig:plot-2prc-d-change}
        }
      \end{minipage}
      \begin{minipage}{0.49\linewidth}
        \centering
        \subfigure[\small{$\theta$=5, $\alpha$=0.5}]{
          {\input{plot/eval_d_5prc.tex}}
          \label{fig:plot-5prc-d-change}
        }
      \end{minipage}
    }
    \caption{Performance of \MTCRP w.r.t. the Latent Dimension $l$}
    \label{fig:d}
  \end{minipage}
  }
\end{figure*}

\subsubsection{Overall Comparison}
\label{sec:experiments:new:overall}
%

\paragraph{\hlgreen{2 Percentile as Sensitivity Threshold}}
\label{sec:experiments:new:overall:2per}

When 2 percentile of the response values (i.e., $\theta$=2) in training data
is used as the sensitivity threshold,
\MTCRP achieves its best AP@$5$ value 0.783, and it is
5.81\% higher than the best AP@$5$ value 0.740 of \BMTMKL ($p$-value=3.565e-8),
\hlgreen{and 3.98\% higher than the best AP@$5$ value 0.753 of {\KRL}
  ($p$-value=9.400e-3). }
In terms of AP@$10$,
\MTCRP achieves its best value 0.758, and it is 6.61\% higher than 
0.711 of \BMTMKL ($p$-value=7.994e-10)
\hlgreen{and 4.55\% higher than the best AP@$10$ value 0.725 of
{\KRL} ($p$-value=6.811e-4).}
Meanwhile, \MTCRP achieves higher AH@$5$ and AH@$10$ compared to those of \BMTMKL
(1.856 vs 1.702, $p$-value=3.379e-11; 2.159 vs 2.072, \mbox{$p$-value=5.805e-8})
\hlgreen{and compared to those of {\KRL} 
(1.856 vs 1.784, $p$-value=2.900e-3;
 2.159 vs 2.137, $p$-value=5.489e-1). }
In particular, \MTCRP achieves its best AP@$k$ and AH@$k$ values
when $\alpha$=0.0, that is, when the
push term $P^{\uparrow}_f$ in Problem~\ref{eqn:obj} is the only
objective to optimize.
%
The results demonstrate that \MTCRP is strong in pushing more sensitive drugs on top of
ranking lists and thus better prioritizes sensitive drugs for drug selection.
On the contrary, \BMTMKL focuses on accurately predicting the response value of each drug in
each cell line. 
However, accurate point-wise response prediction does not guarantee that the most
sensitive drugs are promoted onto the top of ranking
lists in \BMTMKL.
\hlgreen{{\KRL} focuses on optimizing NDCD@$k$ among top-$k$ ranked drugs. However,
  it may heavily rely on the definition of the ``gain'' of a drug at a certain position, and
thus larger gain may not directly lead to more sensitive drugs on top. }

On the other hand, \MTCRP achieves an \rCI value 0.639 when it achieves its best AP@$k$
values (i.e., when $l$=10, $\alpha$=0.0, $\beta$=0.1 and $\gamma$=100.0 for \MTCRP).
Compared to the \rCI value 0.646 of \BMTMKL when \BMTMKL achieves its best AP@$k$ values,
\MTCRP does not outperform \BMTMKL on \rCI. However, the difference is not
significant (-1.08\% increase; $p$-value=3.117e-1).
Note that when $\alpha$=0.0, the ranking orders among sensitive
drugs are not explicitly optimized in Problem~\ref{eqn:obj}. 
\hl{Despite this, {\MTCRP} is able to produce ranking orders}
that are very competitive to those from \BMTMKL.
%
\hlgreen{
  In terms of {\rCI}, the best performance of {\MTCRP} (0.680) and the best 
  performance of {\BMTMKL} (0.646) are both worse than that of {\KRL} (0.683)
  (for {\MTCRP}, -0.44\% increase
  with $p$-value=5.480e-1; for {\BMTMKL}, -5.42\% increase with $p$-value=4.900e-3). 
  Given the low AP@$k$ and AH@$k$ values of {\KRL}, this may indicate that {\KRL} 
  tends to sort sensitive drugs slightly better than {\MTCRP} and {\BMTMKL},
  but does not necessarily rank them on top.
  Different from {\KRL}, {\MTCRP} is able to prioritize sensitive drugs on top while
  maintaining the orders among sensitive drugs reasonably well,
  which is a preferred scenario in drug selection.
}

In addition, \MTCRP achieves a \CI value 0.774 together with its best AP@$k$ values, but
\BMTMKL achieves a \CI value 0.812 with its best AP@$k$ values, which is significantly better
(4.91\% better than \MTCRP, $p$-value=1.966e-123).
As a matter of fact, the best \CI value that \MTCRP ever achieves (i.e., 0.805 when
$l$=30, $\alpha$=0.1, $\beta$=1.0, $\gamma$=100.0) is still significantly worse
than that of \BMTMKL (i.e., 0.812, $p$-value=4.862e-6).
%
%
As \BMTMKL optimizes the predicted response
values, the results indicate that \BMTMKL is able to reproduce
the entire drug ranking structures well using the predicted values.
Different from \BMTMKL, \MTCRP aims to push only sensitive
drugs on top of the ranking structures and optimize only the ranking structures of those
sensitive drugs (when $\alpha > 0$). Therefore, \MTCRP is not able to well estimate the
entire ranking structures for both sensitive and insensitive drugs. 
However, in drug selection, the top ranked drugs could be of great interest compared to
those lower-ranked drugs, and therefore, the low \CI performance of \MTCRP can be
compensated by its high \rCI, AP@$k$ and AH@$k$ values.
\hlgreen{Compared to {\MTCRP} and {\BMTMKL}, {\KRL} achieves the worst {\CI} (0.745).
  Given the fact that {\KRL}
  does not model the ranking relations among insensitive drugs, and considered
  also its high {\rCI} values and median AP@$k$ values, this result may indicate
  that {\KRL} tends to rank some drugs that are very insensitive
  higher than sensitive drugs and higher than the drugs that are less insensitive. }

\paragraph{\hlgreen{5 Percentile as Sensitivity Threshold}}
\label{sec:experiments:new:overall:5per}

%
%
\hl{When 5 percentile of the response values (i.e., $\theta$=5) is used as the sensitivity
threshold, {\MTCRP} shows similar behaviors as in 2 percentile case. 
That is,
in terms of AP@5, {\MTCRP} (0.855 when $l$=10, $\alpha$=0.5, $\beta$=0.1 and $\gamma$=100.0;
0.857 when $l$=50, $\alpha$=0.0, $\beta$=0.1 and $\gamma$=100.0)
outperforms {\BMTMKL} (0.828) at 3.26\% ($p$-value=2.479e-6),
in terms of AP@10 at 4.40\% (0.806 vs 0.772; $p$-value=1.511e-11), and
in terms of AH@5 at 8.37\% (2.965 vs 2.736; $p$-value=2.019e-16) and AH@10 at 5.98\%
(3.986 vs 3.761; $p$-value=2.23e-18).
When {\MTCRP} achieves its best AP@$k$ and AH@$k$,
in terms of {\rCI}, {\MTCRP} achieves similar performance as {\BMTMKL}
(0.658 vs 0.652; $p$-value=2.500e-1), 
and in terms of {\CI}, {\MTCRP} is significantly worse than {\BMTMKL} (0.804 vs 0.812;
$p$-value=3.341e-10).}
%
%
\hlgreen{
  {\KRL} performs similarly at $\theta$=5 as at $\theta$=2, 
  that is, on average, it is worse than {\MTCRP} at AP@$k$, AH@$k$ and {\CI},
  but slightly better at {\rCI} when both achieve their optimal AP@$k$ and AH@$k$.
}

%

%
In particular, the AP@$5$ and AP@$10$ improvement for \MTCRP at $\theta$=2 is larger than that at
$\theta$=5, respectively
(i.e., 5.81\% vs 3.26\% at AP@5, 6.61\% vs 4.40\% at AP@10).
This indicates that \MTCRP is good at prioritizing drugs particularly when there are
a small number of sensitive drugs. 
Note that in Table~\ref{tbl:results:overall_2p5p}, 
for $\theta$=2 and $\theta$=5, the \CI values in \BMTMKL are identical.
This is because \BMTMKL does not use labels in training, and its performance in terms of \CI
does not depend on labels. On the contrary, \rCI depends on the
labels as it only measures \CI within sensitive drugs. Therefore, \rCI values of \BMTMKL
for $\theta$=2 and $\theta$=5 are different.
However, \MTCRP relies on labels during push and ranking in order to learn the models,
and thus, labels will affect its performance in both \CI and \rCI.

%
In Table~\ref{tbl:results:overall_2p5p}, the optimal
\MTCRP results always correspond to non-zero $\gamma$ values
(i.e., the parameter on cell line similarity regularizer in Problem~\ref{eqn:obj}). 
This indicates that cell line similarities calculated from the gene expression information
are able to help improve the ranking of drug sensitivities in \MTCRP.
%
%
The results in Table~\ref{tbl:results:overall_2p5p}
also show that the optimal performance of \MTCRP is from a relatively
small latent space with $l$=10. This may be due to the fact that the sampling scheme
significantly reduces the size of training instances, and thus small latent vectors
are sufficient to represent the learned information for drug prioritization.

\subsubsection{Performance of \MTCRP over Push Powers}
\label{sec:experiments:new:alpha}
%
Fig.~\ref{fig:alpha} presents the best \MTCRP performance on each of the four metrics
with respect to different push parameter $\alpha$'s when $l$=10 (i.e., the 
latent dimension corresponding to the best AP@$k$ values in
Table~\ref{tbl:results:overall_2p5p}).
Fig.~\ref{fig:plot-2prc-alpha-ap-change} and~\ref{fig:plot-5prc-alpha-ap-change}
show that in general as $\alpha$ increases (i.e., decreasing emphasis on pushing sensitive
drugs on top), AP@$k$ values decrease. When $\alpha$=1.0, that is, no push takes effect,
the AP@$k$ values become lower than those when $\alpha < 1$. This demonstrates the effect
of the push mechanism in prioritizing sensitive drugs in \MTCRP.
The figures also show that
%
%
the optimal \rCI values are achieved when $\alpha \in (0, 1)$, but not at $\alpha$=1.0
when the ranking structure among sensitive drugs is the only focus. 
This is probably due to that the ranking difference between
sensitive and insensitive drugs involved in the push term $P^{\uparrow}_f$ can also help
improve the ranking among sensitive drugs. 
In addition, the figures
show that the optimal \CI values are achieved when $\alpha \in (0, 1)$. This is because
with very small $\alpha$ values, sensitive drugs are strongly pushed but it does not
necessarily result in good ranking structures among all sensitive and insensitive
drugs. Similarly, when $\alpha$ is very large, the ranking structures among only
sensitive drugs are highly optimized, which does not necessarily lead to good ranking
structures among all drugs either. Thus, the best overall ranking structures are achieved
under a combinatorial effect of both the push and the sensitive drug ranking.

\subsubsection{Performance of \MTCRP over Latent Dimensions}
\label{sec:experiments:new:l}

%
Fig.~\ref{fig:d} presents the best \MTCRP performance 
on each of the four metrics with respect to different latent dimension $l$. 
Fig.~\ref{fig:plot-2prc-d-change} and~\ref{fig:plot-5prc-d-change} show that in general,
small latent dimensions (e.g., $l$ in 10 to 15) are sufficient in order
to achieve good results on drug ranking. 
%
%
%
%
We interpret each dimension in the drug latent vectors and cell line latent
vectors as to represent a certain latent feature that together determine drug rankings
in each cell line. Thus, the small latent dimensions indicate that the learned
latent vectors are able to capture latent features that are specific to drugs and cell lines.

On the other hand, as AP@$k$ tends to decrease as $l$ increases, \rCI tends to increase.
This indicates that larger $l$ may enable better rankings among sensitive drugs, but not
necessarily pushing sensitive drugs on top. 
Fig.~\ref{fig:d} also shows that \CI first increases and then decreases as $l$ becomes larger,
following an opposite trend of \rCI.
This demonstrates that good ranking structures among all the drugs do not directly
indicate good ranking structures among sensitive drugs, and vice versa.
We also notice that with $\alpha$=0.5,
\MTCRP has better AP@$k$ as $l$ increases from 30 to 50.
This is probably because sufficiently large latent dimensions could also capture 
the drug sensitivity information when the sensitivity threshold is relaxed 
(i.e., more drugs are considered as sensitive when $\theta$=5 than those when $\theta$=2).
Even though, \MTCRP still performs better at $l$=10 than at $l$=50 with $\theta$=5. 
Considering computational costs, we do not explore other even larger latent dimensions.
\begin{table*}[h!]
  \centering
  \caption{{Performance on Ranking New and Known Drugs (\%)}}
  \label{tbl:results:transductive}        
  \begin{small}
  \begin{threeparttable}
      \begin{tabular}{  
        @{\hspace{10pt}}c@{\hspace{10pt}}
        @{\hspace{10pt}}c@{\hspace{10pt}}
        @{\hspace{10pt}}r@{\hspace{10pt}}         
        @{\hspace{10pt}}c@{\hspace{10pt}}
        @{\hspace{10pt}}r@{\hspace{10pt}}
        @{\hspace{10pt}}r@{\hspace{15pt}}
        @{\hspace{10pt}}c@{\hspace{10pt}}
        @{\hspace{10pt}}r@{\hspace{15pt}}
        @{\hspace{10pt}}r@{\hspace{10pt}}
        @{\hspace{10pt}}r@{\hspace{10pt}} 
        @{\hspace{10pt}}r@{\hspace{10pt}}
        @{\hspace{10pt}}r@{\hspace{10pt}}
        }

        \toprule
        sthr &  method   & \multicolumn{6}{c}{parameters} & \multicolumn{4}{c}{performance}\\

        \midrule


        \multirow{3}{*}{-} & \multirow{3}{*}{\BMTMKL}
        & \multicolumn{2}{c}{$\alpha_b$}  & \multicolumn{2}{c}{$\beta_b$} & usim & $\sigma$ & AT@$5$ & AT@$10$ & NT@$5$ & NT@$10$ \\
        \cmidrule(r){3-12}

        & & \multicolumn{2}{c}{1.0e0}   & \multicolumn{2}{c}{1.0e0}   & \hl{$\RBF$} & 10.0 & \textbf{48.03} &	\textbf{54.57} &	47.57           &	54.00 \\
        & & \multicolumn{2}{c}{1.0e-10} & \multicolumn{2}{c}{1.0e-10} & \hl{$\RBF$} & 10.0 & 47.99          &	54.49          &	\underline{\textbf{47.62}}  &	\textbf{54.01} \\

        \midrule

        \multirow{9}{*}{$\theta=2$}

        
        & \multirow{4}{*}{\hlgreen{{\KRL}}}
        & \multicolumn{2}{c}{$k$}  & \multicolumn{2}{c}{$\lambda$} & usim   & $\sigma$ & AT@$5$ & AT@$10$ & NT@$5$ & NT@$10$ \\
        \cmidrule(r){3-12}

        & & \multicolumn{2}{c}{10}   & \multicolumn{2}{c}{1.0e-06}   & $\RBF$ & 0.01   & \underline{\textbf{78.88}} & \underline{\textbf{78.24}} &          3.43      &         28.51  \\
        & & \multicolumn{2}{c}{10}   & \multicolumn{2}{c}{50}        & $\LIN$ & -      &         45.64  &             51.44  & \textbf{45.99}    &         51.81  \\
        & & \multicolumn{2}{c}{10}   & \multicolumn{2}{c}{1.0e-06}   & $\RBF$ & 0.0001 &         45.63  &             51.88  &         45.55     & \textbf{52.28} \\

        \cmidrule(r){2-12}

        & \multirow{5}{*}{\MTCRP}
        & $l$        & $\alpha$ & $\beta$ & $\gamma$ & usim & $\sigma$ & AT@$5$ & AT@$10$ & NT@$5$ & NT@$10$ \\
        \cmidrule(r){3-12}

        & & 50   &   0.50 &   0.1 &    1.0   &  \hl{$\RBF$} &    10.0 &   \textbf{71.38}  &  64.27          &  3.83  &  9.29  \\
        & & 50   &   0.10 &   1.0 &    10.0  &  \hl{$\COS$} &    -    &   62.92           &  \textbf{65.41} &  0.69  &  2.00  \\
        & & 5    &   0.10 &   0.1 &    100.0 &  \hl{$\RBF$} &    10.0 &   49.84 &  58.47 &  \textbf{46.68} &  55.60 \\
        & & 5    &   0.05 &   0.1 &    100.0 &  \hl{$\RBF$} &    10.0 &   49.66 &  58.47 &  46.53 &  \underline{\textbf{55.71}} \\

        \midrule

        \multirow{8}{*}{$\theta=5$}

        & \multirow{3}{*}{\hlgreen{{\KRL}}}
        & \multicolumn{2}{c}{$k$}  & \multicolumn{2}{c}{$\lambda$} & usim   & $\sigma$ & AT@$5$ & AT@$10$ & NT@$5$ & NT@$10$ \\
        \cmidrule(r){3-12}

        & & \multicolumn{2}{c}{10}   & \multicolumn{2}{c}{1.0e-06}   & $\RBF$ & 0.01   & \textbf{75.97} & \textbf{75.07} &         2.64      &          4.14  \\
        & & \multicolumn{2}{c}{10}   & \multicolumn{2}{c}{1.0e-04}   & $\LIN$ & -      &         48.25  &         54.54  & \textbf{46.42}    & \textbf{54.51} \\

        \cmidrule(r){2-12}


        & \multirow{4}{*}{\MTCRP}
        & $l$        & $\alpha$ & $\beta$ & $\gamma$ & usim & $\sigma$ & AT@$5$ & AT@$10$ & NT@$5$ & NT@$10$ \\
        \cmidrule(r){3-12}

        & & 50   &   1.00 &   0.1 &    10.0  &  \hl{$\COS$} &    -    &   \underline{\textbf{79.42}} &  76.49 &  16.99 &  36.06   \\
        & & 50   &   0.50 &   0.1 &    100.0 &  \hl{$\COS$} &    -    &   74.93 &  \underline{\textbf{77.36}} &  5.67  &  17.29   \\
        & & 5    &   0.50 &   0.1 &    10.0  &  \hl{$\RBF$} &    10.0 &   49.84 &  57.97 &  \textbf{45.50} &  \underline{\textbf{54.86}}   \\

        \bottomrule
      \end{tabular}
      \begin{tablenotes}
        \begin{scriptsize}
          \setlength\labelsep{0pt}
        \item
                The columns corresponding to ``$\alpha_b$'', ``$\beta_b$'',
                ``usim'', and ``$\sigma$'' have the
                two hyperparameters, cell line similarity function,
                and parameter for \hl{$\RBF$} cell line similarity, respectively, for \BMTMKL.
                \hlgreen{The columns corresponding to ``$k$'', ``$\lambda$'',
                ``usim'', and ``$\sigma$'' have the
                two hyperparameters, cell line similarity function,
                and parameter for {$\RBF$} cell line similarity, respectively, for {\KRL}.}
                The columns corresponding to ``$l$'', ``$\alpha$'',
                ``$\beta$'', ``$\gamma$'', ``usim'', and ``$\sigma$'' have the
                latent dimension, weighting factor,
                latent vector regularization parameter,
                cell line similarity regularization parameter,
                cell line similarity function, and parameter for \hl{$\RBF$} cell line similarity,
                respectively, for \MTCRP.
          The best performance of each method under each metric is in \textbf{bold}.
	  The best performance of both the methods under each metric is \underline{underscored}.     
          \par
        \end{scriptsize}
      \end{tablenotes}
  \end{threeparttable}
  \end{small}
\vspace{-10pt}
\end{table*}


\subsection{Ranking New and Known Drugs}
\label{sec:experiments:transductive}
%
%
%
We evaluate the performance of \MTCRP on ranking both new drugs (i.e., testing drugs)
and known drugs (i.e., training drugs) together in the experimental setting as in
Fig.~\ref{fig:5fold}.
This corresponds to the use scenario in which new drugs need to be compared with known
drugs so as to select the most promising drugs among all available (i.e., both new and
known) drugs. In this case, we focus on evaluating whether most of the true sensitive drugs
can be prioritized. 
%

%
\subsubsection{Evaluation Metrics}
\label{sec:experiments:transductive:metrics}

The evaluation is based on the following two specific metrics.
The first metric, denoted as AT@$k$, measures among the top-$k$ most sensitive drugs
of each cell line in the ground truth (including both training and testing drugs),
what percentage of them are ranked still among top $k$ in the prediction, that is,
\begin{equation}
  \label{eqn:at@t}
  \text{AT}@k(\{\drug_i\}, \cellline, f)
  = \sum_{\drug_j \in \text{top-$k$}(\cellline)} \frac{\mathbb{I}(\drug_{\overrightarrow{j}} \in \text{top-$k$}(\cellline))}{k}, 
\end{equation}
where $\drug_{\overrightarrow{j}}$ is the drug that is ranked at position $j$ by $f$, and
$\text{top-}k(\cellline)$ is the set of top-$k$ most sensitive drugs in cell line
\cellline. 

The second metric, denoted as NT@$k$, measures among the new drugs that should be among
the top-$k$ most sensitive drugs of each cell line in the ground truth, what percentage of
them are ranked actually among top $k$ in the prediction, that is, 
\begin{equation}
  \label{eqn:nt@t}
  \text{NT}@k(\{\drug_i\}, \cellline, f)
  = \frac{ \displaystyle \sum_{\drug_j \text{ is new}} \mathbb{I}(\drug_{\overrightarrow{j}} \in \text{top-$k$}(\cellline))}{\displaystyle \sum_{\drug_j \text{ is new}} \mathbb{I}(\drug_{j} \in \text{top-$k$}(\cellline))}.
\end{equation}
\hlgreen{%
  The reason why AP$@k$ and AH$@k$ are not used in this experimental setting is that they
  can be easily dominated by known drugs (i.e., training data).
  This is because they are defined for top-$k$ ranked drugs. It is very
  likely that the top-$k$ ranked drugs, when known and new drugs are considered together, 
  are actually known drugs since
  they are explicitly optimized to be on top
  during the training process (for \MTCRP and \KRL).
  %
  Different from AP$@k$ and AH$@k$, AT$@k$ and NT$@k$ are defined
  as particular ratios over the sensitive drugs, which are from both
  training and testing data in this case, and thus less biased by the training data. 
}

\subsubsection{Overall Comparison}
\label{sec:experiments:transductive:overall}
%
Table~\ref{tbl:results:transductive} presents top performance of \BMTMKL, \hl{{\KRL}}
and \MTCRP in terms  of AT@$k$ and NT@$k$.
%
%
\hlgreen{The full set of experimental results is available in
  Table{~\ref{supp:tbl:transductive:2percent:baseline}},
  {\ref{supp:tbl:transductive:5percent:baseline}},
  {\ref{supp:tbl:transductive:2percent:KRL}},
  {\ref{supp:tbl:transductive:5percent:KRL}},
  {\ref{supp:tbl:transductive:2percent}}, and
  {\ref{supp:tbl:transductive:5percent}}
  in the supplementary materials.} 
The results in Table~\ref{tbl:results:transductive} show that in terms of NT@$k$,
\MTCRP is able to 
achieve very similar results (when $l$=5) as \BMTMKL and \hlgreen{\KRL},
in which cases, \MTCRP even achieves slightly better results on AT@$k$ than \BMTMKL
and \hlgreen{{\KRL} when both {\BMTMKL} and {\KRL} achieve their best NT@$k$}. 
This demonstrates that 
\MTCRP has similar power as \BMTMKL \hlgreen{and {\KRL}} in ranking new 
and known sensitive drugs together, and even slightly better power in 
prioritizing new sensitive drugs. 
In terms of AT@$k$, \MTCRP is able to achieve much better results (when $l$=50)
than \BMTMKL. However, when \MTCRP achieves high AT@$k$, the corresponding NT@$k$ is
not optimal.
Since the majority of top-$k$ most sensitive drugs among both new and known drugs
will be known drugs, the good performance of \MTCRP on AT@$k$ validates
that the push mechanism in \MTCRP takes place during training.
\hlgreen{
  In addition, {\MTCRP} outperforms {\KRL} in terms of AT@$k$ when $\theta$=5, but is
  significantly worse than {\KRL} when $\theta$=2.
  This indicates potential overfitting of {\KRL} in ranking sensitive drugs on
  top during training,
  particularly when sensitive training drugs are limited ($\theta$=2) (again, sensitive
  training drugs will be the majority of sensitive drugs). 
}


\subsection{Ranking Drugs in New Cell Lines}
\label{sec:experiments:holdout}

\begin{table*}[h!]
  \centering
  \caption{{Performance on Selecting Drugs for New Cell Lines}}
  \label{tbl:results:holdout}        
  \begin{small}
  \begin{threeparttable}
      \begin{tabular}{  
        @{\hspace{3pt}}c@{\hspace{5pt}}
        @{\hspace{5pt}}c@{\hspace{9pt}}
        @{\hspace{9pt}}r@{\hspace{3pt}}
        @{\hspace{3pt}}r@{\hspace{3pt}}
        @{\hspace{3pt}}r@{\hspace{3pt}}
        @{\hspace{3pt}}r@{\hspace{3pt}}
        @{\hspace{3pt}}r@{\hspace{3pt}}
        @{\hspace{3pt}}r@{\hspace{3pt}}
        @{\hspace{3pt}}r@{\hspace{3pt}}
        @{\hspace{3pt}}r@{\hspace{10pt}}
        @{\hspace{3pt}}r@{\hspace{3pt}}
        @{\hspace{3pt}}r@{\hspace{3pt}}
        @{\hspace{3pt}}r@{\hspace{3pt}}
        @{\hspace{3pt}}r@{\hspace{3pt}}
        @{\hspace{3pt}}r@{\hspace{3pt}}
        @{\hspace{3pt}}r@{\hspace{3pt}}
        @{\hspace{3pt}}r@{\hspace{3pt}}
        @{\hspace{3pt}}r@{\hspace{3pt}}
        }

        \toprule
        sthr & method   & \multicolumn{8}{c}{AP@$5$} & \multicolumn{8}{c}{AP@$10$}\\

        \midrule

        & & 50 & 100 & 150 & 200 & 250 & 300 & 350 & 400 & 50 & 100 & 150 & 200 & 250 & 300 & 350 & 400 \\
        \cmidrule(r){3-10}
        \cmidrule(r){11-18}
        \multirow{3}{*}{$\theta=2$}

        & \BMTMKL
        & 0.855 & 0.842 & 0.844 & 0.834 & 0.833 & 0.829 & 0.823 & 0.829 
        & 0.792 & \textbf{0.783} & 0.783 & \textbf{0.778} & 0.774 & 0.768 & 0.763 & 0.768 \\

        & \hlgreen{\KRL}
        & 0.451 & 0.409 & 0.399 & 0.432 & 0.430 & 0.388 & 0.397 & 0.382 & 0.436 & 0.403 & 0.390 & 0.398 & 0.404 & 0.379 & 0.377 & 0.356 \\

        & \MTCRP
        & \textbf{0.876} & \textbf{0.870} & \textbf{0.870} & \textbf{0.852} & \textbf{0.861} & \textbf{0.856} & \textbf{0.848} & \textbf{0.853} 
        & \textbf{0.800} & 0.770 & \textbf{0.789} & 0.777 & \textbf{0.798} & \textbf{0.780} & \textbf{0.771} & \textbf{0.792} \\

        \midrule

        & & 50 & 100 & 150 & 200 & 250 & 300 & 350 & 400 & 50 & 100 & 150 & 200 & 250 & 300 & 350 & 400 \\
        \cmidrule(r){3-10}
        \cmidrule(r){11-18}
        \multirow{3}{*}{$\theta=5$}

        & \BMTMKL
        & 0.951 & 0.945 & 0.946 & 0.937 & 0.940 & 0.937 & 0.935 & 0.938 
        & 0.903 & \textbf{0.908} & \textbf{0.911} & \textbf{0.907} & \textbf{0.908} & \textbf{0.906} & \textbf{0.903} & \textbf{0.906} \\

        & \hlgreen{\KRL}
        & 0.621 & 0.568 & 0.565 & 0.545 & 0.565 & 0.570 & 0.549 & 0.543 & 0.573 & 0.497 & 0.524 & 0.501 & 0.520 & 0.503 & 0.515 & 0.502 \\

        & \MTCRP
        & \textbf{0.965} & \textbf{0.958} & \textbf{0.955} & \textbf{0.949} & \textbf{0.949} & \textbf{0.949} & \textbf{0.947} & \textbf{0.947}
        & \textbf{0.908} & \textbf{0.908} & 0.908 & 0.906 & 0.904 & \textbf{0.906} & 0.901 & \textbf{0.906} \\

        \bottomrule
      \end{tabular}
      \begin{tablenotes}
        \begin{scriptsize}
          \setlength\labelsep{0pt}
        \item
        The columns corresponding to ``AP@$5$'' and ``AP@$10$'' have the
        AP@$5$ and AP@$10$ evaluation values of the corresponding methods, 
        respectively.
        The columns corresponding to ``50'' $\cdots$ ``400'' have the number of
        new cell lines in the experiment (i.e., $\holdoutsize$).
        The best performance of the three methods under each metric is in 
        \textbf{bold}.
          \par
        \end{scriptsize}
      \end{tablenotes}
  \end{threeparttable}
  \end{small}
\vspace{-10pt}
\end{table*}


In this section, we present the experimental results on ranking drugs in new cell lines.
The experiments follow the experimental setting as in Fig.~\ref{fig:loo}.

\subsubsection{Analysis on Cell Line Similarities}
\label{sec:experiments:holdout:validation}
%

%
New cell lines don't have any drug response information or latent vectors, and the only
information that can be leveraged in order to select drugs for them is their own genomics
information. Therefore, we first validate whether we can use the gene expression information
for drug selection in new cell lines in \MTCRP.  

We first calculate the similarities of cell lines using their latent vectors
learned from \MTCRP (in the setting of Fig.~\ref{fig:5fold}) in \hl{$\RBF$} function.
The correlation between such similarities and the cell line
similarities calculated from gene expressions (i.e., $w_{pq}$ as in Equation~\ref{eqn:sim})
using \hl{$\RBF$} function is 0.426.
The correlations show that cell line gene expression similarities and their latent vector 
similarities are moderately correlated.
%
%
We further analyze the cell lines whose gene expression similarities (using \hl{$\RBF$} function)
are among 90 percentile. For each of such cell lines, we identify 10 most similar cell lines
in their gene expressions. 
Fig.~\ref{supp:fig:sim-compare} in the supplementary materials
shows the gene expression similarities of all such cell lines and
their latent vector similarities.
%
Fig.~\ref{supp:fig:sim-compare} in the supplementary materials
demonstrates that for
those cell lines whose gene expression similarities are high, their latent vector
similarities are also significantly higher than average (the average
  cell line latent vector similarity is 0.682).
This indicates the feasibility of using high gene expression similarities to connect
new cell lines with cell lines used in \MTCRP

\subsubsection{Experimental Setting}
\label{sec:experiments:holdout:setting}

%
Based on the analysis on cell line similarities, we split testing cell lines
(i.e., new cell lines) from training cell lines (as in Fig.~\ref{fig:loo})
such that each of the testing cell lines has sufficient number of similar training cell lines
in terms of their gene expressions. Cell line latent vectors
are learned in \MTCRP only for those training cell lines, and drug latent vectors are
learned for all the drugs. Note that the label scheme in this setting follows that in
Section~\ref{sec:materials:dataset:labelling:new}. 
The detailed protocol is available in Section~\ref{supp:sec:holdout:protocol} in 
supplementary materials.

In order to select sensitive drugs for each of the testing/new cell lines, we first
generate a latent vector for the testing cell line as the weighted sum of latent vectors of
its top-10 most similar (in gene expressions) training cell lines. The weights are
the respective gene expression similarities. The drugs are then scored using the
latent vector of the new cell line and latent vectors of all drugs.

%
%
\subsubsection{Overall Comparison}
\label{sec:experiments:holdout:compare}

%
\hlgreen{Table.~{\ref{tbl:results:holdout}} presents the performance of
  {\BMTMKL}, {\KRL} and {\MTCRP}}
with respect to different numbers of new cell lines, \hl{denoted as {\holdoutsize}}, 
in terms of AP@$5$, AP@$10$, respectively. We don't present the performance in
\rCI and \CI here because in drug selection for new cell lines/patients, \CI
is not practically as indicative as AP@$k$, particularly in drug selection
from a large collection of drugs. 
For each of the two evaluation metrics, we compare the performance of \BMTMKL,
\hlgreen{{\KRL}} and \MTCRP when $\theta$=2 and $\theta$=5. 
Note that as \holdoutsize increases (i.e., more new cell lines), the average gene
expression similarities between new cell lines and training cell lines decrease
according to the data split protocol. 

%
\hlgreen{Table.~{\ref{tbl:results:holdout}}} shows that as \holdoutsize increases, 
the AP@$5$ and AP@$10$ values of 
\hlgreen{{\MTCRP}, {\KRL} and {\MTCRP}}
with both $\theta$=2 and $\theta$=5 decrease in general.
This is because as more cell lines are split into testing set, on average, training cell
lines and testing cell lines are less similar, and thus it is less accurate to extrapolate
from training cell lines to new cell lines. 
%
Even though, \MTCRP consistently outperforms \BMTMKL 
\hlgreen{and {\KRL}} over all \holdoutsize values in terms of AP@$5$. 
and performs similarly to \BMTMKL \hlgreen{and still outperforms \KRL} in terms of AP@$10$. 
Specifically, when 50 cell lines are held out for testing (i.e., $\holdoutsize$=50),
\MTCRP achieves AP@$5$ = 0.876/0.965 when $\theta$ = 2/5, compared to 
AP@$5$ = 0.855/0.951 of \BMTMKL 
\hlgreen{and AP@$5$ = 0.451/0.621 of {\KRL}}.
%
When 400 cell lines are held out for testing, \MTCRP achieves AP@$5$ = 0.853/0.947,
compared to AP@$5$ = 0.829/0.938 of \BMTMKL
\hlgreen{and AP@$5$ = 0.382/0.543 of {\KRL} when $\theta$=2/5}. 
%
Particularly, with $\theta$=2, \MTCRP outperforms \BMTMKL at 2.5\% when $\holdoutsize$=50, 
and at 2.9\% when $\holdoutsize$=400.
\hl{This indicates that when the drug selection for
new cell lines is more difficult} (e.g., fewer training cell lines, fewer sensitive
drugs), \MTCRP outperforms \BMTMKL more. 
Also with $\theta$=2, when more cell lines are held out ($\holdoutsize \ge 250$),
\MTCRP outperforms \BMTMKL in AP@$10$. For example, when $\holdoutsize$=250,
\MTCRP achieves AP@$10$ = 0.798, compared to AP@$10$ = 0.774 of \BMTMKL. 
This also indicates that \MTCRP outperforms \MTCRP on more difficult drug selection
problems. 
\hlgreen{
  Please note that {\KRL} performs significantly worse than {\MTCRP} and {\BMTMKL}
  in selecting drugs for new cell lines. 
  This might be due to similar reasons as discussed in 
  Section{~\ref{sec:experiments:transductive:overall}}, 
  that is, {\KRL} tends to overfit the known (training) cell lines 
  and is not well generalizable to new cell lines.
}
\subsection{Analysis on Latent Vectors}
\label{sec:experiments:lf}
%

\subsubsection{Analysis on Drug Latent Vectors}
\label{sec:experiments:lf:drugs}

\paragraph{Evaluation Measurements}
\label{sec:experiments:lf:drugs:measurements}

%
We evaluate how much the learned drug latent vectors 
could be interpreted in differentiating
sensitive drugs and insensitive drugs. \hl{To have quantitative measurements for such
an evaluation}, we calculate the following four types of measurement: 
\begin{enumerate}[noitemsep,nolistsep,leftmargin=*]
\item the cosine similarities of drugs using their latent vectors learned from \MTCRP,
denoted as \siml; 
\item the Tanimoto coefficients~\cite{Wale2008} 
  of drugs using their
AF features~\footnote{\url{http://glaros.dtc.umn.edu/gkhome/afgen/overview}},
denoted as \simt; 
\item the average ranking percentile difference for all the drug pairs over all the cell
  lines in the ground truth, denoted as \prdiff; and
\item the average difference of responsive cell line ratios for drug pairs 
  over all the cell lines in the ground truth, denoted as \efdiff.
\end{enumerate}
AF features are binary fingerprints representing whether a certain substructure
is present or not in a drug. Thus, the Tanimoto coefficients over AF features
measure how drugs are similar in terms of their intrinsic structures (Tanimoto coefficient
has been demonstrated to be effective in comparing drug structures~\cite{Wale2008}). 
The measurement \prdiff is calculated on all pairs of drugs
over the cell lines that both of the drugs in a pair have sensitivity measurement (i.e., no
missing values on either of the drugs) in the cell lines.
The absolute values of the percentile ranking
differences over such cell lines are then averaged into \prdiff. 
The measurement \efdiff is
calculated as the percentage of cell lines in which a drug is sensitive
(with $\theta$=5).
The absolute values of such ratio differences from all the drug pairs \mbox{are then
averaged into \efdiff}.

\paragraph{Discriminant Power of Drug Latent Vectors}
\label{sec:experiments:lf:drugs:power}
%
%
\begin{figure}[!h]
\vspace{-10pt}
  \begin{center}
    \scalebox{0.9}{
      \input{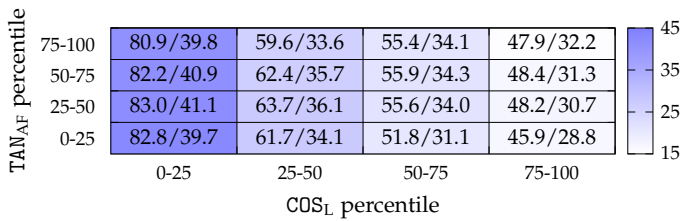}
    }
    \caption{\prdiff in different drug pairs}
    \label{fig:grid_prank}
  \end{center}
\vspace{-10pt}
\end{figure}
We group all the drug pairs based on their \siml and \simt percentile values. 
Fig.~\ref{fig:grid_prank} presents the \prdiff for
different groups of drug pairs.
In Fig.~\ref{fig:grid_prank}, the colors code the \prdiff values. The two
values in each drug group (e.g., $x/y$ in each cell in the figure)
are the average percentile ranking of
the higher-ranked drugs (i.e., $x$) and of the lower-ranked drugs (i.e., $y$)
in the drug pairs, respectively. The difference of the two values in each drug
group is the corresponding \prdiff.
Fig.~\ref{fig:grid_prank} shows that when the drugs are less similar in their latent
vectors (i.e., smaller \siml percentile; the left columns in Fig.~\ref{fig:grid_prank}),
the drugs are ranked more differently among cell lines on average (i.e., larger \prdiff
values). 
%
\hl{
Drugs with similar latent vectors tend to be ranked similarly (i.e., 
smaller {\prdiff} values in the right columns in {Fig.~\ref{fig:grid_prank}} with larger {\siml}
percentiles).
}
This indicates that the drug latent vectors learned from \MTCRP are able to encode
information that differentiates drug rankings in cell lines.

\begin{figure}[!h]
\vspace{-15pt}
  \begin{center}
    \scalebox{0.9}{
      \input{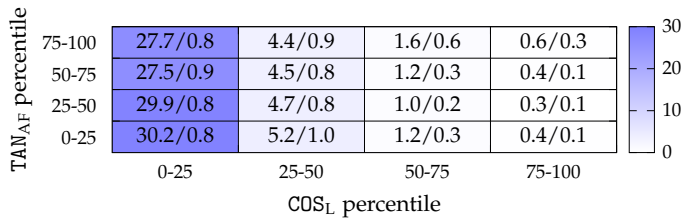}
    }
    \caption{\efdiff in different drug pairs}
    \label{fig:grid_top5}
  \end{center}
\vspace{-15pt}
\end{figure}
Fig.~\ref{fig:grid_top5} presents the \efdiff for different groups of drug pairs. 
In Fig.~\ref{fig:grid_top5}, the colors code the \efdiff values. The two
values in each drug group (e.g., $x/y$) are the average responsive cell line ratio
of the higher-ranked drugs (i.e., $x$) and of the lower-ranked drugs (i.e., $y$)
in the drug pairs, respectively. The difference of the two values in each drug
group is the corresponding \efdiff.
%
%
Fig.~\ref{fig:grid_top5} shows that drugs that are very different from others
in \siml (i.e., smaller \siml percentile; the left columns in Fig.~\ref{fig:grid_top5})
are sensitive in more cell lines (i.e., larger $x$ in $x/y$ values of the left
columns).
Specifically, the higher-ranked drugs (i.e., corresponding to $x$ in $x/y$ values in
Fig.~\ref{fig:grid_top5}) in the 4 ranges of \siml values (in increasing order)
are sensitive in 28.8\%, 4.7\%, 1.3\% and 0.4\% of the cell lines on average, respectively. 
\hl{This also corresponds with what Fig.~{\ref{fig:grid_prank}} shows,}
that is,
drugs that are more different from others in \siml tend to be ranked
higher than the drugs that are more similar to others in \siml.
Specifically, in Fig.~\ref{fig:grid_prank},
the higher-ranked drugs (i.e., corresponding to $x$ in $x/y$ values in
Fig.~\ref{fig:grid_prank}) in the 4 ranges of \siml values (in increasing order)
are ranked at 82.2, 61.9, 54.7 and 47.6 percentile on average, respectively.
These indicate that the sensitive drugs are better differentiated in drug latent
vectors, and thus \MTCRP is effective in deriving drug latent vectors that are
specific to drug sensitivities.

\paragraph{\hl{Drug Latent Vectors as New Drug Features}}
\label{sec:experiments:lf:drugs:features}
\hl{
Analysis on drug latent vectors as new drug features is available in
Section{~\ref{sec:drugs:features}} in the supplementary materials. }

\paragraph{\hlgreen{Interpretability of Drug Latent Vectors}}
\label{sec:experiments:lf:drugs:intepret}
\hlgreen{
Analysis on the interpretability of drug latent vectors is available in
Section{~\ref{sec:drugs:intepret}} in the supplementary materials. }

\subsubsection{Analysis on Cell Line Latent Vectors}
\label{sec:experiments:lf:cell}
%
%
\begin{figure*}[!h]
  \begin{center}
    \hspace{28pt}
    \input{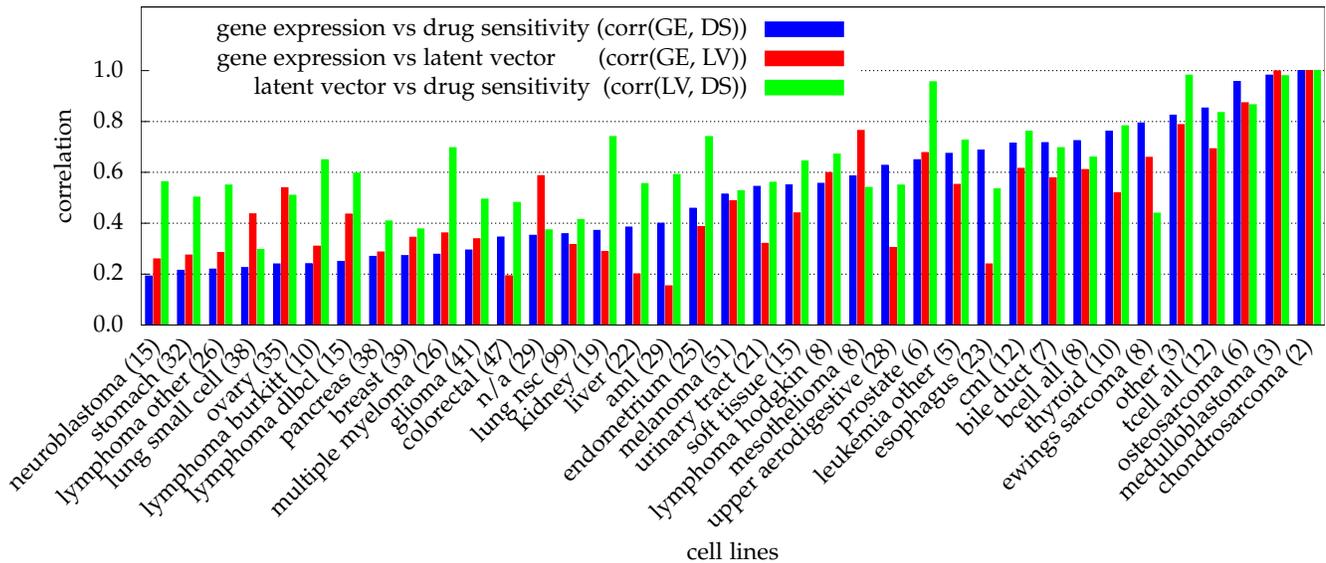}
    \vspace{-40pt}
      \caption{Correlation among Different Cell Line Similarities}
    \label{fig:sim-corr-spearman}
  \end{center}
\vspace{-20pt}
\end{figure*}
Fig.~\ref{fig:sim-corr-spearman} presents the correlations among three different types of
cell line similarities within each of the tumor types.
The three cell line similarities are calculated from 
gene expressions (GE) using \hl{$\RBF$} function \hl{on the 1,203 selected genes}, 
cell line latent vectors (LV) using \hl{$\RBF$} function
and drug sensitivity profiles (DS) using Spearman rank correlation coefficient.
The three corresponding correlations are denoted as \gl, \gd and \ld, respectively. 
The numbers associated with tumor types in Fig.~\ref{fig:sim-corr-spearman} indicate the
number of cell lines of corresponding tumor types. 
(e.g., melanoma (51) indicates that there are 51 cell lines of melanoma).
Among the 37 tumor types as originally categorized in CCLE,
28 tumor types (i.e., 75.7\% of all tumor types)
have their \ld higher than or same as \gd, and the average percentage difference is
59.9\%. For example, for 15 neuroblastoma cell lines, \ld is on average 191.7\% higher
than \gd. For all the cell lines of various lymphoma, \ld is on average at least 20\%
higher than \gd. 
This indicates that even when the correlation between gene expression and
drug sensitivity is not strong, through learning cell line latent vectors, \MTCRP can
discover novel cell line features (i.e., cell line latent vectors) that better
characterize their drug response patterns.
As a matter of fact, the improvement of \ld over \gd is more significant when \gd is lower
(i.e., the left side of the panel in Fig.~\ref{fig:sim-corr-spearman}). 
\hl{This indicates {\MTCRP} may be effective for cell lines which do not have significant correlation
between selected gene expression and drug sensitivity profiles.}
For the cell lines whose \gd is large (i.e., the right side of the panel in
Fig.~\ref{fig:sim-corr-spearman}), \ld is still high in general and meanwhile \gl is also
high.
%
\hl{This indicates that {\MTCRP} may be effective in retaining the useful signals from gene expression
  when the correlation between selected gene expression and drug sensitivity profiles is strong.}
For a few tumor types with relatively low \gl (e.g., liver, \hl{AML} and esophagus), 
their \ld is actually relatively high. This may indicate the capability of \MTCRP
in learning new signals for cell lines by leveraging information \hl{from} multiple
other cell lines. 

%

\section{Discussions and Conclusions}
\label{sec:conclusion}

We developed genomics-regularized joint push and learning-to-rank method \MTCRP to tackle
cancer drug selection for three particular application scenarios: 1). select sensitive
drugs from new drugs for each known cell line;
2). select sensitive drugs from all available drugs including new and known drugs for
each known cell line; 
and 3). select sensitive drugs \hl{from} all available drugs for new cell lines.
Our new method \MTCRP outperforms or achieve similar performance to the 
state-of-the-art methods \BMTMKL and \KRL. 
\hlgreen{
  Note that {\BMTMKL} was originally developed to select cell lines for drugs. However,
  {\BMTMKL} uses kernel regression and approximates drug response values to sort cell lines.
  Such drug response values can also be used to select drugs for cell lines, as we used
  in this manuscript.
  {\KRL} was originally developed to select drugs for cell lines -- the same
  goal as our {\MTCRP}, but shows low generalization
  capabilities in selecting drugs for new cell lines and from new drugs.}

\hlgreen{
  {\MTCRP} is a pair-wise learning-to-rank method which scores each drug in each cell
  line using latent vectors and ranks drugs based on the latent scores.
  This is fundamentally different from {\BMTMKL} (a point-wise learning-to-rank method)
  and {\KRL} (a list-wise learning-to-rank method), in
  which the drug response values are approximated and used to sort drugs.
  The latent scores enable more flexibility to learn accurate drug ranking structures,
  but fall short when drug response values are also of concerns.
  We will investigate a combination of latent scoring and drug response value
  regression to leverage the advantages of both the schemes. 
}

\hl{In {\MTCRP}, each drug has a global latent vector which is the same in all cell lines.}
This might be restrictive as the learned drug latent vectors may have to compromise
their performance in some cell lines in order to achieve better performance in
other cell lines, and thus better overall performance. We will explore personalized
drug latent vectors in the future research, that is, each drug will have different
latent vectors with respect to different cell lines. In this way, the ranking
performance on each cell line is expected to be further improved. 

\hlgreen{In addition, in {\MTCRP} and {\BMTMKL},
  gene selection has been done across all tissue types/origins
  for all the cell lines together. 
  However, different tissue types/origins
  may have different genes useful in their drug rankings. 
  Therefore, global gene selection may end up with genes which tend to be globally
  indicative for assessing drug sensitivities, while locally informative gene
  information has been neglected.
  We will explore localized gene selection in the future research,
  by clustering cell lines based on tissues/origins and conducting gene selection
  over each cluster. We will also explore cell line clustering with respect to
  their drug ranking structures. 
  It is expected that by localized gene selection, drug selection could be more
  accurate for each cell line. 
}

\hlgreen{The {\MTCRP} method suffers from the lack of interpretability.
  It is not easy to explain what each of
  the drug/cell line latent dimensions represents or
  whether they correlate to any mechanisms of actions.
  This is because the latent vectors are learned only with respect to the goal of
  optimizing the drug ranking structures, without any potential
  MOA (if ever known) or causal features (if ever known) explicitly modeled.
  We would tackle this challenging interpretability issue in the future research,
  for example, by integrating known knowledge in the latent vector learning. 
}

We will also evaluate our \MTCRP method on other drug-cell line screening data,
for example, NCI60~\footnote{https://dtp.cancer.gov/discovery\_development/nci-60/} and LINCS-L1000~\footnote{http://www.lincsproject.org/LINCS/} data. When the number of
drugs (chemical compounds in LINCS-L1000) is large, it becomes more challenging
computationally when pairs of drugs are used in learning. We will explore fast
learning algorithms to learn drug latent vectors in the future research.

%

\ifCLASSOPTIONcompsoc
  \section*{Acknowledgments}
\else
  \section*{Acknowledgment}
\fi
  This material is based upon work supported by the National Science Foundation
  under Grant Number IIS-1566219 and IIS-1622526. 

\bibliographystyle{IEEEtran}
\bibliography{paper,ds,ds2}

%


\vspace{-20pt}
\begin{IEEEbiography}[{\includegraphics[width=1in,height=1.25in,clip]{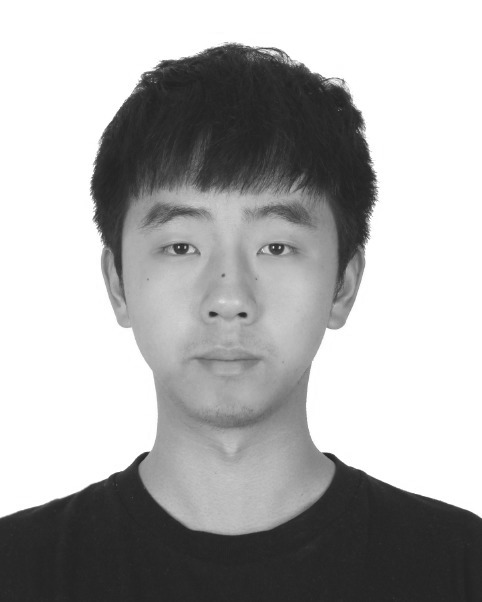}}]{Yicheng He}
  received his B.S. degrees from the Department of Computer \& Information Science,
  Indiana University-Purdue University, Indianapolis, Indiana, and 
  from School of Software, Sun Yat-sen University, China, in 2016.
  His research is on data mining and machine learning with applications in drug discovery. 
\end{IEEEbiography}

\vspace{-20pt}
\begin{IEEEbiography}[{\includegraphics[width=1in,height=1.25in,clip]{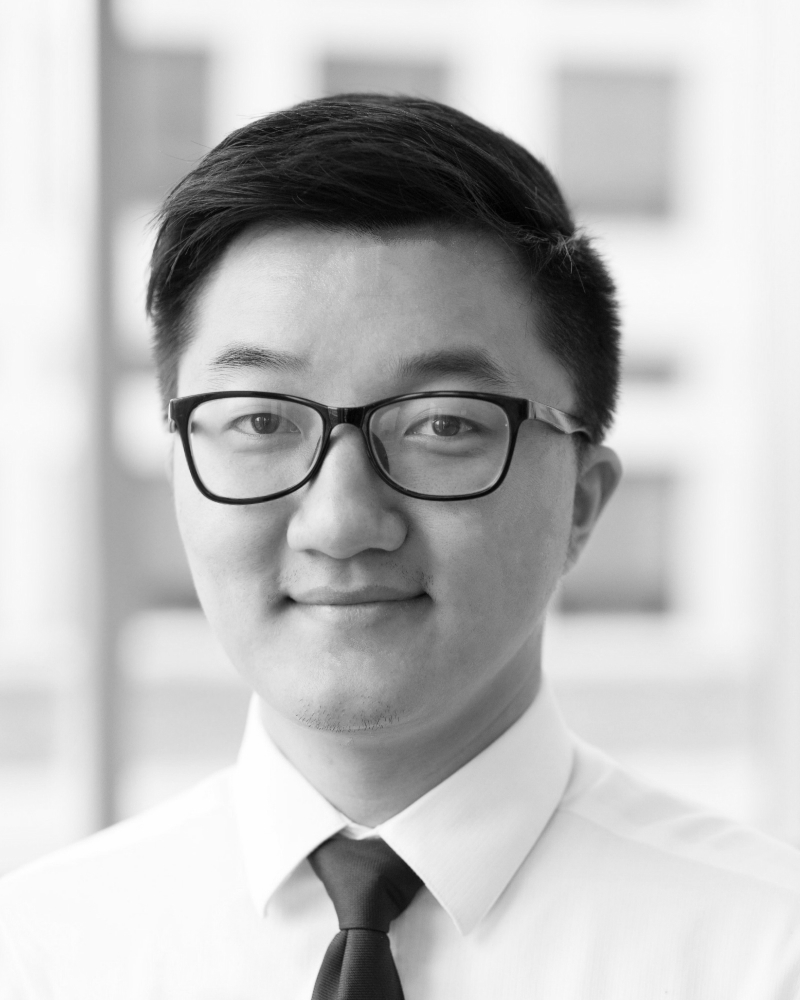}}]{Junfeng Liu}
  received his M.S. degree from the Department of Computer \& Information Science, 
  Indiana University-Purdue University, Indianapolis, Indiana, in 2018. 
  He is currently a Graduate Research Assistant at Indiana University-Purdue University Indianapolis. 
  His research is on data mining and machine learning with applications in drug discovery, 
  bioinformatics and healthcare informatics.
\end{IEEEbiography}


\vspace{-20pt}
\begin{IEEEbiography}[{\includegraphics[width=1in,height=1.25in,clip]{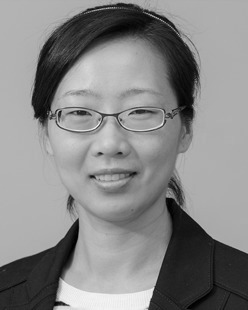}}]{Xia Ning}
  received her Ph.D. degree from the Department of Computer Science \& Engineering,
  University of Minnesota, Twin Cities, USA, in 2012.
  She is currently an Assistant Professor at the Department of Computer
  \& Information Science, Indiana University-Purdue University, Indianapolis, Indiana. Her
  research is on data mining and machine learning with applications in drug discovery,
  bioinformatics and healthcare informatics. 
\end{IEEEbiography}

\newpage



\vfill


\end{document}